\documentstyle[jair,twoside,11pt,theapa]{article}
\input epsf 

\jairheading{31}{2008}{113-155}{08/07}{01/08}
\ShortHeadings{CTL Model Update for System Modifications} {Zhang \& Ding}

\long\def\comment#1{}

\newtheorem{examp}{Example} 
\newenvironment{example}{\begin{examp}\rm}{$\Box$\end{examp}}
\newtheorem{definition}{Definition} 
\newtheorem{lemma}{Lemma} 
\newtheorem{proposition}{Proposition} 

\newtheorem{theorem}{Theorem}
\newenvironment{proof}{{\bf Proof:}}{$\Box$\\ }
\long\def\comment#1{}

\begin{document}

\title{CTL Model Update for System Modifications}


\author{\name Yan Zhang \email yan@scm.uws.edu.au\\
 \addr Intelligent Systems Laboratory\\
School of Computing and Mathematics\\
University of Western Sydney, Australia
\AND
\name Yulin Ding \email yulin@cs.adelaide.edu.au\\
\addr Department of Computer Science\\
University of Adelaide, Australia
}


\maketitle

\begin{abstract}
Model checking is a promising technology, which has been applied
for verification of many hardware and software systems. In this
paper, we introduce the concept of {\em model update} towards the
development of an automatic system modification tool that extends
model checking functions. We define primitive update operations on
the models of Computation Tree Logic (CTL) and formalize the principle
of minimal change for CTL model update. These primitive update
operations, together with the underlying minimal change principle,
serve as the foundation for CTL model update. Essential semantic
and computational characterizations are provided for our CTL
model update approach. We then describe a formal algorithm that
implements this approach. We also illustrate
two case studies of CTL model
updates for the well-known microwave oven example 
and the Andrew File System 1, 
from which we further propose
a method to optimize the update results in complex system modifications.

\comment{
Minimal change is a fundamental principle for modeling system
dynamics. In this paper, we study the issue of minimal change for
Computational Tree Logic (CTL) model update. We first propose five
primitive operations which capture the basic update
of the CTL model, and then define
the minimal change criteria for CTL model update based on these primitive
operations.
We provide essential semantic and computational characterizations for
our CTL model update approach. We develop a formal algorithm to implement
this update that employs the underlying minimal change principle. We
also present a CTL model update example using
the well known Microwave oven scenario.\\
}
%
\end{abstract}

\section{Introduction}

Model checking is one of the most effective technologies for
automatic system verifications. In the model checking approach, 
the system behaviours are modeled by 
a Kripke structure,
and specification properties that we require the system to meet are expressed
as formulas in a propositional temporal logic, e.g., CTL.
Then the model checker, e.g., SMV, takes the Kripke model and a formula as input, 
and verifies whether the formula is satisfied by the Kripke model.
If the formula is not satisfied in the Kripke model,
the system will report errors, and possibly provides useful information
(e.g., counterexamples). 
 
Over the past decade, the model checking technology has been 
considerably developed, and many effective model checking tools
have been demonstrated
through provision of thorough automatic error diagnosis in complex
designs e.g.,
\cite{Amla&etal05,Berard&etal01,Boyer&Sighireanu03,Chauhan&etal02,wing95}.
Some current state-of-the-art model checkers, such as
SMV~\cite{Clarke&etal99}, NuSMV~\cite{Cimatti&etal99} and Cadence
SMV~\cite{McMillan&Amla02}, employ SMV specification language for
both Computational Tree Logic (CTL) and Linear Temporal Logic (LTL)
variants~\cite{Clarke&etal99,HuthandRyan2000}. Other model checkers,
such as SPIN~\cite{Holzmann03}, use Promela specification language
for on-the-fly LTL model checking. Additionally, the
MCK~\cite{Gammie&Meyden04} model checker was developed by
integrating a knowledge operator into CTL model checking to verify
knowledge-related properties of security protocols.

\comment{
Using the model checking concept, a system which is to be checked by
the model checker is represented in a Kripke model, and
specification properties are represented as specific logic formulas
e.g., CTL formulas. Then, the model checker (e.g., SMV) takes the Kripke
model and a formula as input, and verifies whether the latter is
satisfied by the former. The system will report errors, and possibly
provide useful information (e.g., counterexamples), if the formula
is not satisfied in the Kripke model.
}

Although model checking approaches 
have been used for verification of
problems in large complex systems, one major limitation of
these approaches is that they can only verify the correctness of a
system specification. In other words, if errors are identified in
a system specification by model checking, the task of correcting
the system is completely left to the system designers. That is,
model checking is generally used only to verify the correctness of
a system, not to modify it. Although the idea of repair has been
indeed proposed for model-based diagnosis, repairing a system is
only possible for specific
cases~\cite{Dennis&etal06,Stumptner&Wotawa96}.

\subsection{Motivation}

Since model checking can handle complex system verification
problems and as it may be implemented via fast algorithms, it is
quite natural to consider whether we can develop associated
algorithms so that they can handle system modification as well.
The idea of integrating model checking and automatic modification
has been investigated in recent years. 
Buccafurri, Eiter, Gottlob, and Leone \citeyear{Buccafurri99} 
have proposed an approach whereby AI
techniques are combined with model checking such that the enhanced
algorithm can not only identify errors for a concurrent system,
but also provide possible modifications for the system.

In the above approach, a system is described as a Kripke structure
$M$, and a modification $\Gamma$ for $M$ is a set of state transitions that
may be added to or removed from $M$. If a CTL formula $\varphi$ is
not satisfied in $M$ i.e., the system contains errors with respect
to property $\varphi$, then $M$ will be repaired by adding new
state transitions or removing existing ones specified in $\Gamma$. As a
result, the new Kripke structure $M'$ will then satisfy formula
$\varphi$.
The approach of Buccafurri et al. \citeyear{Buccafurri99} 
integrates model checking and
abductive theory revision to perform system repairs.
They also demonstrate how their
approach can be applied to repair concurrent programs.

It has been observed that this type of system repair is quite
restricted, as only relation elements (i.e., state transitions)
in a Kripke model can be
changed\footnote{NB: No state changes occur in the specified
system repairs (see Definitions 3.2 and 3.3 in
\citeR{Buccafurri99}).}. This implies that errors can only be fixed
by changing system behaviors. In fact, as we will show in this
paper, allowing change to both states and relation elements in a Kripke
structure significantly enhances the system repair process in
most situations. Also, since providing all admissible modifications
(i.e., the set $\Gamma$) is a pre-condition of any repair, the
approach of Buccafurri et al. lacks flexibility.
Indeed, as stated by the authors themselves,
their approach may not be general enough for other system
modifications.

On the other hand, knowledge-base update has been the subject of
extensive study in the AI community since the late 1980s.
Winslett's Possible Model Approach (PMA) is viewed as pioneering
work towards a model-based minimal change approach for
knowledge-base update \cite{Winslett88}. Many researchers have
since proposed different approaches to knowledge system update
(e.g., see references from 
Eiter \& Gottlob, 1992; Herzig \& Rifi, 1999).
Of these works, Harris and Ryan~\citeyear{Harris&Ryan02,Harris&Ryan03}
considered using an update approach for system modification, where
they designed update operations to tackle feature integration,
performing theory change and belief revision. However, their study
focused mainly on the theoretical properties of system update, and
practical implementation of their approach in system modification
remains unclear.

Baral and Zhang~\citeyear{Baral&Zhang03} recently developed a formal
approach to knowledge update based on single-agent S5 Kripke
structures observing that system modification is closely related
to knowledge update. From the knowledge dynamics perspective, we
can view the finite transition system, which represents a real
time complex system, to be a model of a knowledge set (i.e., a
Kripke model). Thus the problem of system modification is reduced
to the problem of updating this model so that a new updated model
satisfies the knowledge formula.

This observation motivated the initial development of a general
approach to updating Kripke models, which can be integrated into
model checking technology, towards a more general automatic system
modification. Ding and Zhang's work \citeyear{Ding&Zhang05} may be
viewed as the first attempt to apply this idea to LTL model update.
The LTL model update modifies the existing LTL model of an
abstracted system to automatically correct the errors occurring
within this model.

Based on the investigation described above, we intend to integrate
knowledge update and CTL model checking to develop a practical model updater, which
represents a general method for automatic system repairs.

\subsection{Contributions of This Paper}

\begin{figure}[tbhp]
\begin{center}
\epsfysize = 40mm
\epsffile{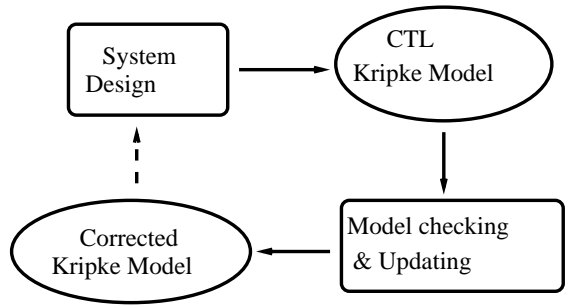}
\caption{CTL model update.}
\label{fig11}
\end{center}
\end{figure}

The overall aim of our work  is to design a model updater that
improves model checking function by adding error repair (see
schematic in Figure~\ref{fig11}). The outcome from the updater is a
corrected Kripke model. The model updater's function is to
automatically correct errors reported (possibly as counterexamples) by a model
checking compiler. Eventually, the model updater is intended to be a
universal compiler that can be used in certain common situations for
model error detection and correction.

The main contributions of this paper are described as follows:
\begin{enumerate}
\item We propose a formal framework for
CTL model update. Firstly, we define primitive CTL model
update operations and, based on these operations, specify a minimal
change principle for the CTL model update. We then study the relationship between 
the proposed CTL model update and traditional propositional belief update.
Interestingly, we prove that our CTL model update obeys all Katsuno and Mendelzon
update postulates (U1) - (U8).
We further provide important
characterizations for special CTL model update formulas
such as $\mbox{EX}\phi$, $\mbox{AG}\phi$ and $\mbox{EG}\phi$.
These characterizations play an
important role in optimization of the update procedure. Finally, we 
study the computational properties of CTL model update and show
that, in general, the model checking problem for CTL model update is
co-NP-complete. We also classify a useful subclass of CTL model update problems
that can be performed in polynomial time.

\item We develop a formal algorithm for CTL model update.
In principle, our algorithm can perform an update on a given CTL Kripke model
with an arbitrary satisfiable CTL formula and generate a model that
satisfies the input formula and has a minimal change with respect to the original model.
The model then can be viewed as a possible 
correction on the original system specification.
Based on this algorithm, we implement a system prototype of CTL model updater
in C code in Linux.

\item We demonstrate important applications of
our CTL model update approach
by two case studies of the well-known microwave oven example 
\cite{Clarke&etal99} and the Andrew File System 1
\cite{wing95}. Through these case studies, we further propose a new update principle of
minimal change with maximal reachable states, which can significantly improve the update results
in complex system modification scenarios.
\end{enumerate}

In summary, our work presented in this paper is an initial step 
towards the formal study of the automatic 
system modification. This approach may
be integrated into existing model checkers
so that we may develop a unified methodology and system for model checking 
and model correction. In this sense, our work will enhance the 
current model checking technology.
Some results presented
in this paper were published in ECAI 2006 \cite{yulin-yan06}.

The rest of the paper is organized as follows. An overview of CTL
syntax and semantics is provided in Section~\ref{sec:CTL}. 
Primitive update operations on CTL models are defined in
Section~\ref{sec:MinimalChange}, and a minimal change principle
for CTL model update is then developed.
Section~\ref{sec:characterizations} consists of a study of the relationship between
CTL model update and Katsuno and Mendelzon's update postulates
(U1) - (U8), and various 
characterizations for some special CTL model updates. In
Section~\ref{sec:ComputationalProp},  a general
computational complexity result of CTL model update is proved, and
a useful tractable subclass of CTL model update problems is identified.
A formal algorithm for the proposed CTL model update approach is
described in Section~\ref{sec:Algorithms}. 
In Section 7, 
two update
case studies are illustrated to demonstrate applications of our CTL model 
update approach. Section~\ref{ExplosionID} proposes an improved 
CTL model update approach which can significantly optimize the update results
in complex system modification scenarios.  Finally, the
paper concludes with some future work discussions in
Section~\ref{sec:Conclusions}.

\section{Preliminaries}

In this section, we briefly review the syntax and semantics of
Computation Tree Logic and basic concepts of belief update, which are
the foundation for our CTL model update.

\subsection{CTL Syntax and Semantics}
\label{sec:CTL}

To begin with, we briefly review CTL syntax and semantics (refer to
\citeR{Clarke&etal99} and \citeR{HuthandRyan2000} for details).

\begin{definition}
\label{def:CTL-Kripke}
Let $AP$ be a set of atomic propositions. A Kripke model $M$ over AP
is a triple $M=(S,R,L)$ where:
\begin{enumerate}
\item $S$ is a finite set of states;
\item $R\subseteq S\times S$ is a binary relation representing state transitions;
\item $L:S\rightarrow 2^{AP}$ is a labeling function that assigns each state
with a set of atomic propositions.
\end{enumerate}
\end{definition}

\begin{figure}[tbhp]
\begin{center}
\epsfysize = 40 mm
\epsffile{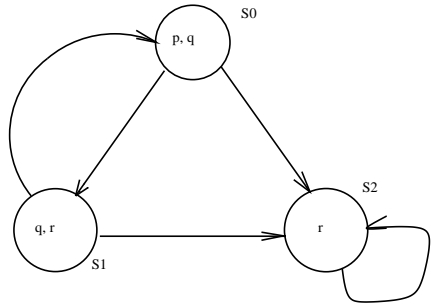}
\caption{Transition state graph.} 
\label{f:KripkeModel}
\end{center}
\end{figure}

An example of a finite Kripke model is represented by the graph in
Figure~\ref{f:KripkeModel}, where each node represents a state in
$S$, which is attached to a set of propositional atoms being assigned by the 
labeling function, and an edge represents a state transition
- a relation element in $R$ describing a system
transition from one state to another.  

\comment{ In
Figure~\ref{f:KripkeModel}, state $s_0$ contains propositional atoms
$p$ and $q$ being true; state $s_1$ contains $q$ and $r$ being true
and state $s_2$ containing $r$ being true. These propositional atoms
are $L(s)$ in Definition~\ref{def:CTL-Kripke}. 
}

Computation Tree Logic (CTL) is a temporal logic 
allowing us to refer to the future. It is also a branching-time
logic, meaning that its model of time is a tree-like structure in
which the future is not determined but consists of different paths,
any one of which might be the `actual' path that is eventually
realized~\cite{HuthandRyan2000}.

\begin{definition}
\label{def:LTL-syntax}
CTL has the following syntax given in Backus-Naur form:

\vspace*{0.1in} $\phi::=\top \mid \perp \mid p \mid (\neg \phi) \mid (\phi_1 \wedge
\phi_2)\mid (\phi_1 \vee \phi_2) \mid \phi \rightarrow \psi \mid \mbox{\em AX} \phi \mid
\mbox{\em EX} \phi$

 \hspace*{0.4in} $\mid \mbox{\em AG}\phi\mid \mbox{\em EG}\phi\mid \mbox{\em AF}\phi
\mid \mbox{\em EF}\phi\mid \mbox{\em A}[\phi_1 \mbox{\em U} \phi_2]\mid 
\mbox{\em E}[\phi_1 \mbox{\em U} \phi_2]$

\vspace*{0.1in} where $p$ is any propositional atom.
\end{definition}

A CTL formula is evaluated on a Kripke model. A path in a Kripke model from
a state is a(n) (infinite) sequence of states.
Note that for a given path, the same state may occur an infinite number of
times in the
path (i.e., the path contains a loop). To simplify our following discussions,
we may identify states in a path with different position subscripts,
although states occurring in different positions in the path may be the same.
In this way, we can say that one state precedes another in a path without
much confusion. Now we can present useful notions in a formal way.
Let $M=(S,R,L)$ be a Kripke model and $s\in S$. A {\em path} in $M$
starting from $s$ is denoted as 
$\pi= [s_0, s_1, \cdots,s_{i-1},s_i,s_{i+1},\cdots]$, where $s_0=s$ and
$(s_i,s_{i+1})\in R$ holds for all $i\geq 0$. We write $s_i\in \pi$
if $s_i$ is a state occurring in the path $\pi$.  If a path
$\pi=[s_0,s_1,\cdots,s_i,\cdots,s_j,\cdots]$ and $i<j$, we also 
denote $s_i < s_j$. Furthermore for a given path $\pi$, we use notion
$s\leq s_i$ to denote a state $s$ that is the state $s_i$ or $s<s_i$.
  For simplicity, we may use $succ(s)$ to
denote state $s'$ if there is a relation element $(s,s')$ in $R$.

\begin{definition}
\label{def:CTL-semantic}
Let $M =(S,R,L)$ be a Kripke model for CTL.
Given any $s$ in $S$, we define whether a CTL formula $\phi$ holds
in $M$ at state $s$. We denote this by $(M, s)\models \phi$. The
satisfaction relation $\models$ is defined by
structural induction on all CTL formulas:
\begin{enumerate}
\item  $(M, s) \models \top$ and $(M,s)\not \models \perp$ for all
$s\in S$.
\item  $(M,s)\models p$ iff $p\in L(s)$.
\item $(M,s)\models \neg \phi$ iff $(M,s) \not\models \phi$.
\item $(M,s)\models\phi_1\wedge\phi_2$ iff $(M,s)\models\phi_1$ and
$(M,s)\models\phi_2$.
\item $(M,s)\models\phi_1\vee\phi_2$ iff
$(M,s)\models\phi_1$ or $(M,s)\models\phi_2$.
\item $(M,s)\models
\phi_1\rightarrow \phi_2$ iff $(M,s)\models \neg\phi_1$, or
$(M,s)\models \phi_2$.
\item $(M,s)\models \mbox{\em AX}\phi$ iff for all $s_1$
such that $(s,s_1)\in R$, $(M,s_1)\models \phi$.
\item  $(M,s)\models \mbox{\em EX}\phi$ iff for some $s_1$ such that
$(s,s_1)\in R$, $(M,s_1)\models\phi$.
\item  $(M,s)\models \mbox{\em AG}\phi$  iff for all
paths $\pi=[s_0, s_1, s_2, \cdots]$ where $s_0=s$ and  $\forall
s_i,~s_i\in\pi$,  $(M,s_i)\models \phi$.
\item  $(M,s)\models \mbox{\em EG}\phi$  iff there is
a path $\pi=[s_0, s_1, s_2, \cdots]$ where $s_0=s$ and $\forall
s_i,~s_i\in \pi$,  $(M,s_i)\models \phi$.
\item $(M,s)\models \mbox{\em AF}\phi$  iff for all
paths $\pi=[s_0, s_1, s_2, \cdots]$ where $s_0=s$ and  $\exists
s_i,~s_i\in \pi$, $(M,s_i)\models \phi$.
\item $(M,s)\models \mbox{\em EF}\phi$ iff there is a path
$\pi=[s_0,s_1, s_2, \cdots]$ where $s_0 = s$ and $\exists
s_i,~s_i\in\pi$, $(M,s_i)\models \phi$.
\item $(M,s)\models \mbox{\em A}[\phi_1\mbox{\em U}\phi_2]$  iff for all paths
$\pi=[s_0, s_1, s_2, \cdots]$ where $s_0=s$, 
$\exists s_i\in\pi$, $(M,s_i)\models
\phi_2$ and for each $j<i$,  $(M,s_j)\models \phi_1$.
\item $(M,s)\models \mbox{\em E}[\phi_1\mbox{\em U}\phi_2]$  iff there is a path
$\pi=[s_0, s_1, s_2, \cdots]$ where $s_0=s$, 
$\exists s_i\in \pi$, $(M,s_i)\models
\phi_2$ and for each $j<i$, $(M,s_j)\models \phi_1$.
\end{enumerate}
\end{definition}

From the above definition, we can see that the intuitive meaning of 
A, E, X, and G are quite clear: A means for all paths, E means that there exists a path,
X refers to the next state and G means for all states globally.
Then the semantics of a CTL formula is easy to capture
as follows.

In the first six clauses, the truth value of the formula in the
state depends on the truth value of $\phi_1$ or $\phi_2$ in the same
state. For example, the truth value of $\neg\phi$ in a state only
depends on the truth value of $\phi$ in the same state. This
contrasts with clauses $7$ and $8$ for AX and EX. For instance, the
truth value of AX$\phi$ in a state $s$ is determined not by $\phi$'s
truth value in $s$, but by $\phi$'s truth values in states $s'$
where $(s,s')\in R$; if $(s, s)\in R$, then this value also depends
on the truth value of $\phi$ in $s$.

The next four clauses ($9$ - $12$) also exhibit this phenomenon.
For example, the truth value of AG$\phi$ involves looking at the
truth value of $\phi$ not only in the immediately related states,
but in indirectly related states as well. In the case of AG$\phi$,
we must examine the truth value of $\phi$ in every state related
by any number of forward links (paths) to the current state $s$.
In clauses $13$ and $14$,  symbol $\mbox{U}$ may be explained as ``until'':
a path $\pi=[s_0,s_1,s_2, \cdots]$ satisfies $\phi_1\mbox{U}\phi_2$ if
there is a state $s_i\in\pi$ such that for all $s<s_i$, 
$(M,s)\models \phi_1$ {\em until} $(M,s_i)\models\phi_2$.

\begin{figure}[tbhp]
\begin{center}
\epsfysize = 60mm 
\epsffile{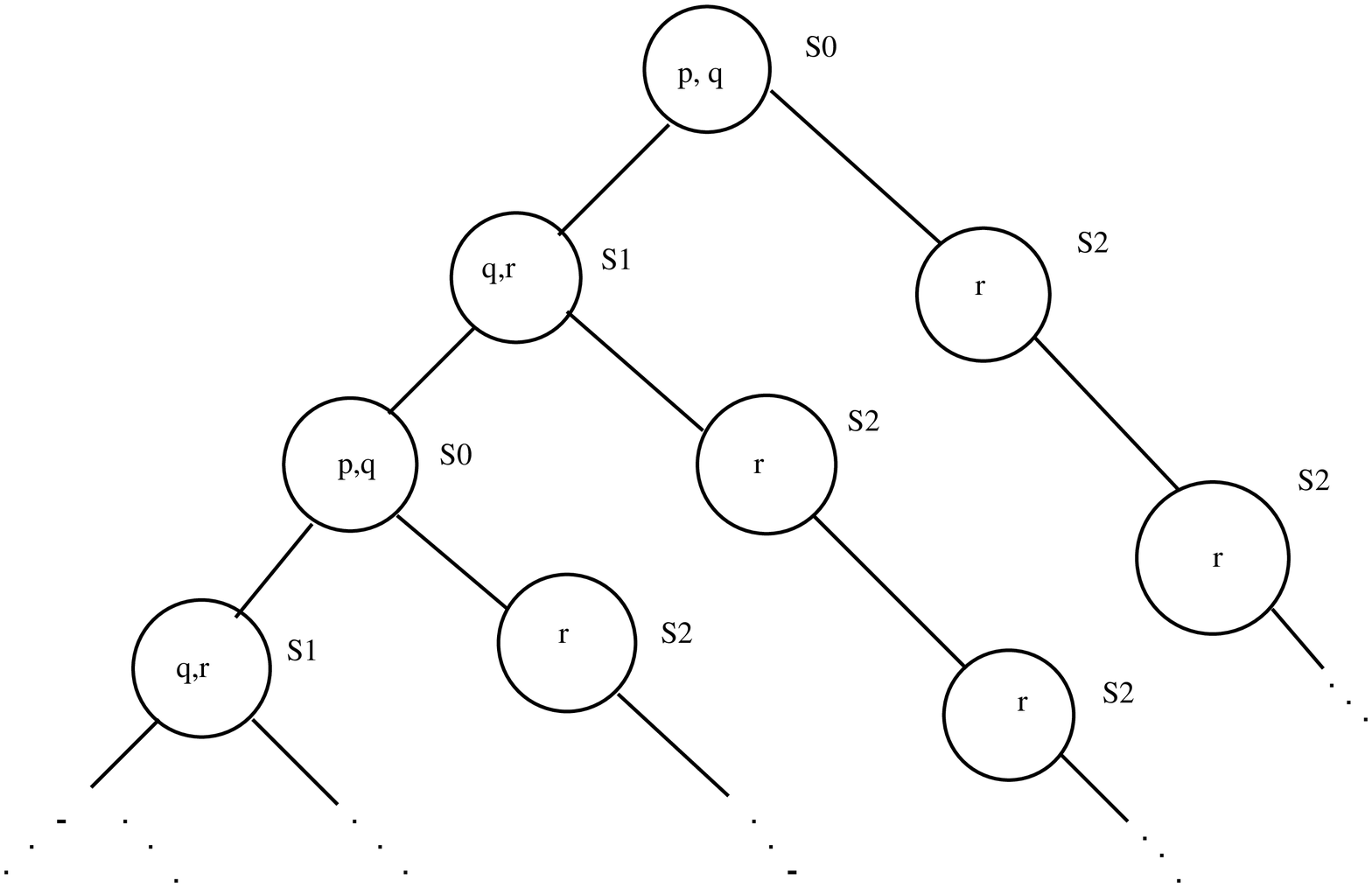}
\caption{Unwinding the transition state graph as an infinite tree.}
\label{f:UnwoundTree}
\end{center}
\end{figure}

Clauses $9$ - $14$ above refer to computation paths in models. It is,
therefore, useful to visualize all possible computation paths from a
given state $s$ by unwinding the transition system to obtain an
infinite computation tree. This greatly facilitates deciding whether
a state satisfies a CTL formula. The unwound tree of the graph in
Figure~\ref{f:KripkeModel} is depicted in
Figure~\ref{f:UnwoundTree} (note that we assume $s_0$ is the initial state in 
this Kripke model).

\begin{figure}[tbhp]
\begin{center}
\epsfysize = 60 mm
\epsffile{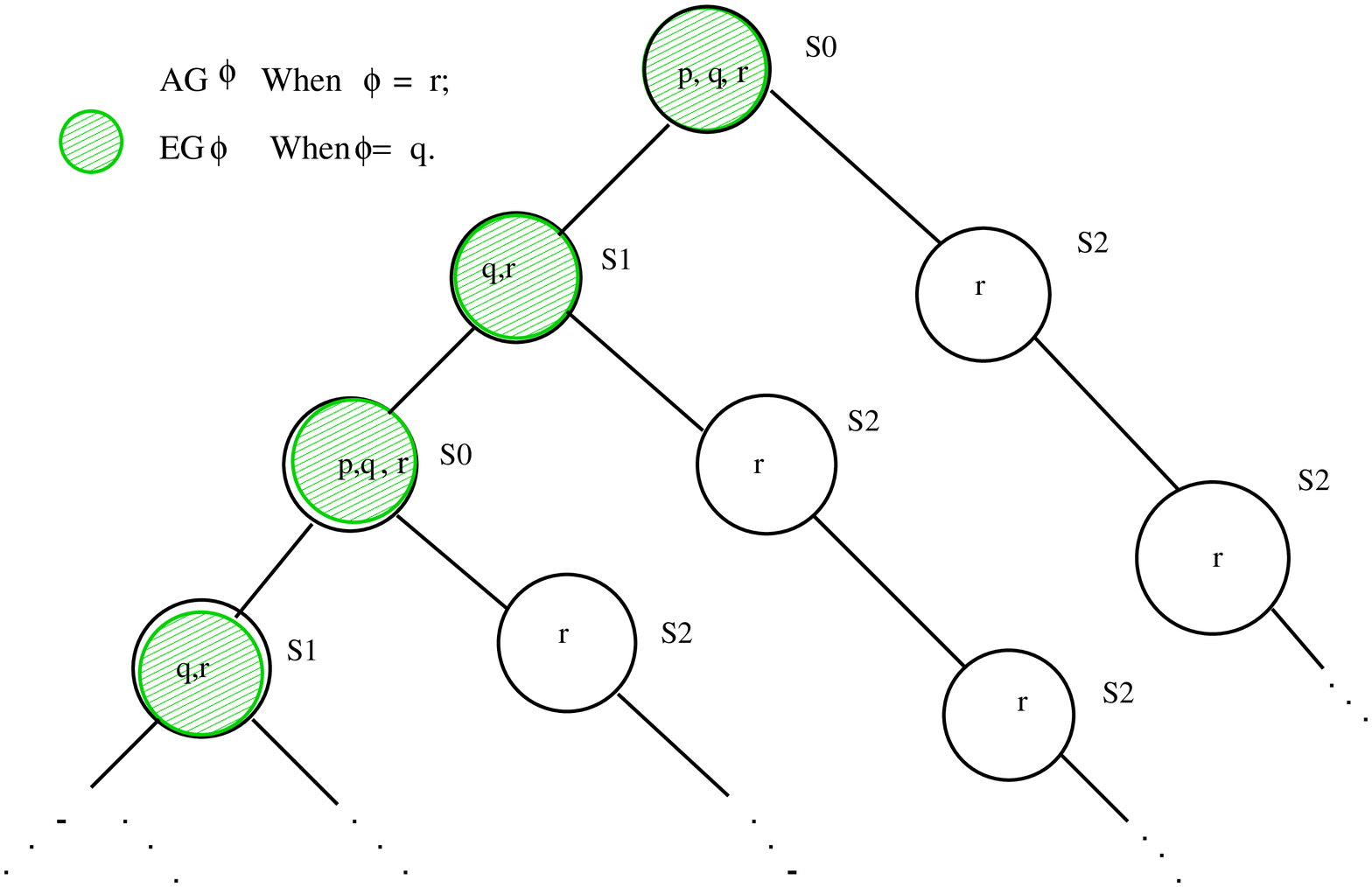}
\caption{AG$\phi$ and EG$\phi$ in an unwound tree.}
\label{f:AEG}
\end{center}
\end{figure}

In Figure~\ref{f:UnwoundTree}, if $\phi =r$, then AX$r$ is true;
if $\phi =q$, then EX$q$ is true. In the same figure, if $\phi
=r$, then AF$r$ is true because some states on all paths will
satisfy $r$ some time in the future. If $\phi =q$, EF$q$ is true
because some states on some paths will satisfy $q$ some time in
the future. 
The clauses for
AG and EG can be explained in Figure~\ref{f:AEG}. In this tree,
all states satisfy $r$. Thus, AG$r$ is true in this Kripke model.
There is one path where all states satisfy $\phi =q$. Thus, EG$q$
is true in this Kripke model.

The following De Morgan rules and 
equivalences \cite{HuthandRyan2000} will be useful for our CTL model update algorithm 
implementation: 

\[\neg \mbox{AF}\phi \equiv \mbox{EG}\neg\phi ; \]
\[\neg \mbox{EF}\phi \equiv \mbox{AG}\neg\phi ; \]
\[\neg \mbox{AX}\phi \equiv \mbox{EX}\neg \phi ;\]
\[\mbox{AF}\phi \equiv \mbox{A}[\top \mbox{U} \phi] ;\]
\[\mbox{EF}\phi \equiv \mbox{E}[\top \mbox{U} \phi] ;\]
\[\mbox{A}[\phi_1\mbox{U} \phi_2] \equiv\neg (\mbox{E}[\neg \phi_2\mbox{U} (\neg\phi_1\wedge\phi_2)]
\vee \mbox{EG}\neg \phi_2). \]

In the rest of this paper, without explicit declaration, we will
assume that all CTL formulas occurring in our context will be
satisfiable. For instance, if we consider updating a Kripke model to
satisfy a CTL formula $\phi$, we already assume that
$\phi$ is satisfiable.

From Definition \ref{def:CTL-semantic}, 
we can see that for a given CTL Kripke model $M=(S,R,L)$, if $(M,s)\models \phi$ and
$\phi$ is a propositional formula, then $\phi$'s truth value solely 
depends on the labeling function $L$'s assignment on state $s$. In this case
we may simply write $L(s)\models \phi$ if there is no confusion from the context.

\subsection{Belief Update}

Belief change has been a primary research topic in the AI community 
for almost two decades e.g., \cite{agm,Winslett90}. Basically, it studies 
the problem of how an agent can change its beliefs when it wants to bring
new beliefs into its belief set. There are two types of belief changes, namely 
{\em belief revision} and {\em belief update}.  Intuitively, belief revision 
is used to modify a belief set in order to 
accept new information about the static world, while
belief update is to bring the belief set up to date when the world is described by
its changes.

Katsuno and Mendelzon \citeyear{kat92} have discovered that 
the original AGM revision postulates cannot precisely characterize 
the feature of belief update. They 
proposed the following alternative update postulates, and argued that any propositional
belief update operators should satisfy these postulates. 
In the following (U1) - (U8) postulates, all occurrences of
$T$, $\mu$, $\alpha$, etc. are propositional
formulas.

\begin{quote}
 (U1)\hspace{.05in} $T\diamond\mu \models \mu$.\\
 (U2)\hspace{.05in} If $T \models \mu$ then $T\diamond\mu\equiv T$.\\
 (U3)\hspace{.05in} If both $T$ and $\mu$ are satisfiable
then $T\diamond\mu$ is also satisfiable.\\
 (U4)\hspace{.05in} If
$T_{1}\equiv T_{2}$ and $\mu_{1}\equiv\mu_{2}$ then
$T\diamond\mu_{1}\equiv T_{2}\diamond\mu_{2}$.\\
 (U5)\hspace{.05in}
$(T\diamond\mu )\wedge\alpha \models T\diamond (\mu\wedge\alpha )$.\\
 (U6)\hspace{.05in} If
$T\diamond\mu_{1} \models \mu_{2}$ and
$T\diamond\mu_{2} \models \mu_{1}$ then $T\diamond\mu_{1}\equiv
T\diamond\mu_{2}$.\\
 (U7)\hspace{.05in} If $T$ is complete (i.e.,
has a unique model) then\\
\hspace*{.3in} 
$(T\diamond\mu_{1})\wedge (T\diamond\mu_{2}) \models T\diamond (\mu_{1}\vee\mu_{2})$.\\
 (U8)\hspace{.05in} $(T_{1}\vee T_{2})\diamond\mu\equiv$
$(T_{1}\diamond\mu)\vee (T_{2}\diamond\mu )$.
\end{quote}

As shown by Katsuno and Mendelzon \citeyear{kat92}, postulates (U1) - (U8) precisely 
capture the minimal change criterion for update that is defined
based on certain partial ordering on models. As a typical model based belief update
approach, here we briefly introduce Winslett's
Possible Models Approach (PMA) \cite{Winslett90}.
We consider a propositional language ${\mathcal L}$.
Let $I_1$ and $I_2$ be two Herband interpretations of ${\mathcal L}$. The {\em 
symmetric difference} between
$I_1$ and $I_2$ is defined as $diff(I_1,I_2)=(I_1 - I_2)\cup (I_2 -I_1)$. Then for a given
interpretation $I$, we define a partial ordering 
$\leq_{I}$ as follows: $I_1\leq_{I} I_2$ if and only if
$diff(I,I_1)\subseteq diff(I,I_2)$. Let ${\mathcal I}$ be a collection of interpretations,
we denote $Min({\mathcal I},\leq_{M})$ to be the set of all minimal models from
${\mathcal I}$ with respect to ordering $\leq_{M}$, where
model $M$ is fixed.  Now
let $\phi$ and $\mu$ be two propositional formulas, the update of $\phi$ with $\mu$ using
the PMA, denoted as $\phi\diamond_{pma}\mu$, is defined as follows:
\begin{quote}
$Mod(\phi\diamond_{pma}\mu)=\bigcup_{M\in Mod(\phi)} Min(Mod(\mu),\leq_{M})$,
\end{quote}
where $Mod(\psi)$ denotes the set of all 
models of formula $\psi$.
It can be proved that the PMA update operator $\diamond_{pma}$
satisfies all postulates (U1) - (U8). 

Our work of CTL model update has a close connection to the idea of 
belief update. As will be
shown in this paper, in our approach, 
we view a CTL Kripke model as a description of the world that we are interested in,
i.e., the description of a system of dynamic behaviours, and the update on this Kripke 
model occurs when the setting of the system of dynamic behaviours has to change
to accommodate some desired properties.
Although there is a significant difference between classical propositional belief update
and our CTL model update, we will show that Katsuno Mendelzon's update postulates
(U1) - (U8) are also suitable to characterize the minimal change principle for our CTL model
update.

\section{Minimal Change for CTL Model Update}
\label{sec:MinimalChange}

We would like to extend the idea of minimal change in belief update to our CTL model 
update. In principle, when we need to update a CTL Kripke model to satisfy
a CTL formula, we expect the updated model to retain as much information as possible 
represented in the original model. In other words, 
we prefer to change the model in a minimal way to achieve our goal.  
In this section, we will propose formal metrics of minimal change for CTL model 
update.

\subsection{Primitive Update Operations}

Given a CTL Kripke model and a (satisfiable) CTL formula, we
consider how this model can be updated in order to satisfy the given
formula. From the discussion in the previous section,  
we try to incorporate a minimal change principle into our update approach. 
As the first step towards this aim, we should have a way to measure the difference between
two CTL Kripke models in relation to a given model.
We first illustrate our initial consideration of this aspect through an example.

\comment{
\begin{definition}
\label{def:update} ({\bf CTL Model Update}) Given a CTL Kripke
model $M=(S,R,L)$ and a satisfiable CTL formula $\phi$, an {\em update} of
${\mathcal M}=(M,s_0)$ to satisfy $\phi$, where $s_0\in S$, is a
CTL Kripke model $M'=(S',R',L')$, such that
$(M',s_{0}')\models \phi$ where $s_{0}'\in S'$ and 
$M'$ is a minimal model with respect to ordering $\leq_{M}$.
We use $Update({\mathcal M},\phi)$ to denote the result ${\mathcal M'}=(M,s_0')$
\end{definition}
}

\begin{example}
Consider a simple CTL model
$M=(\{s_0,s_1,s_2\}, \{(s_0,s_0)$, $(s_0,s_1)$, $(s_0,s_2)$, $(s_1,s_1)$, $(s_2,s_2)$,
$(s_2,s_1)\}, L)$, where  
$L(s_0)=\{p,q\}, L(s_1)=\{q,r\}$ and $L(s_2)=\{r\}$.
$M$ is described as in Figure \ref{ex1}.

\begin{figure}[tbhp]
\begin{center}
\epsfysize = 40 mm
\epsffile{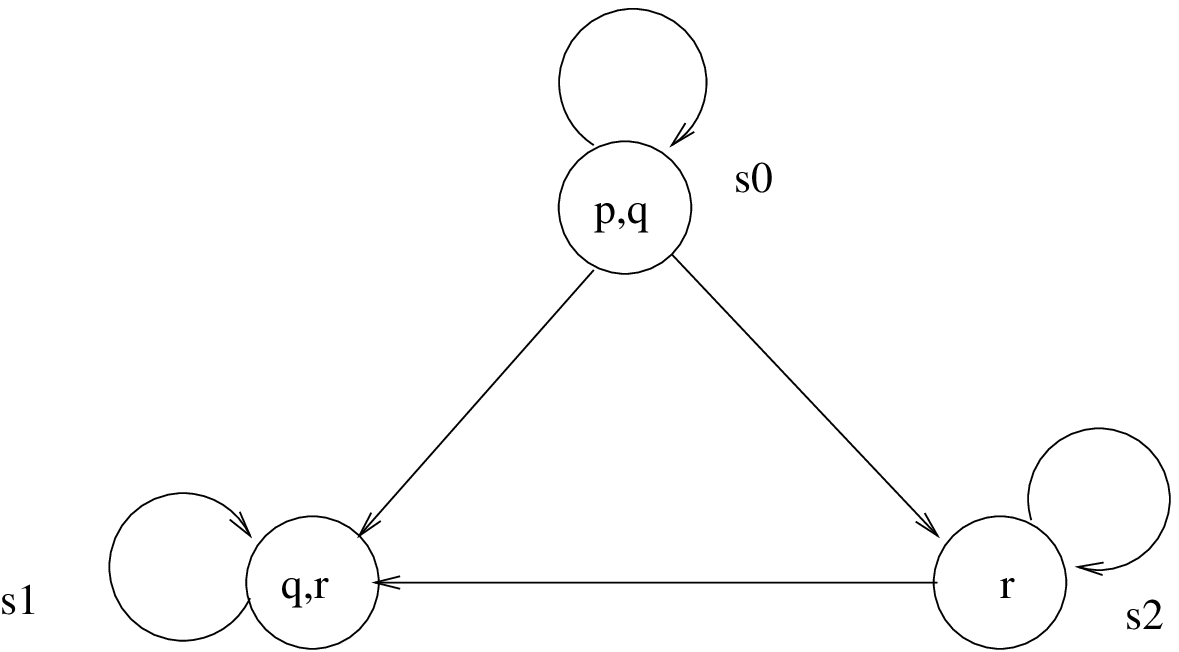}
\caption{Model $M$.}
\label{ex1}
\end{center}
\end{figure}

Now consider formula AG$p$. Clearly, $(M,s_0)\not\models$ AG$p$.
One way to update $M$ to satisfy AG$p$ is to update states $s_1$
and $s_2$ so that both updated states satisfy $p$\footnote{Precisely,
we update the labeling function $L$ that changes the truth assignments to
$s_1$ and $s_2$.}. Therefore, we
obtain a new CTL model $M'=(\{s_0,s_1,s_2\},
\{(s_0,s_0),(s_0,s_1),(s_0,s_2)$, $(s_1,s_1)$, $(s_2,s_2)$,
$(s_2,s_1)\}$, $L')$, where $L'(s_0)=L(s_0)=\{p,q\},
L'(s_1)=\{p,q,r\}$ and $L'(s_2)=\{p,r\}$. In this update, we can see that
the labeling function has been changed to associate different truth 
assignments with states $s_1$ and $s_2$.
Another way to update $M$
to satisfy formula AG$p$ is to simply remove relation elements $(s_0,s_1)$
and $(s_0,s_2)$ from $M$, this gives $(M'', s_0)\models$ AG$p$,
where $M''=(\{s_0,s_1,s_2\}, \{(s_0,s_0), (s_1,s_1), (s_2,s_2),
(s_2,s_1)\}, L)$. This more closely resembles the approach of
Buccafurri et al. ~\cite{Buccafurri99}, where no state changes
occur. It is interesting to note that the first of the updated
models retains the same ``structure'' as the original, while it is
significantly changed in the second. These two possible results
are described in Figure~\ref{ex1-1}.
\end{example}

\begin{figure}[tbhp]
\begin{center}
\epsfysize = 35 mm
\epsffile{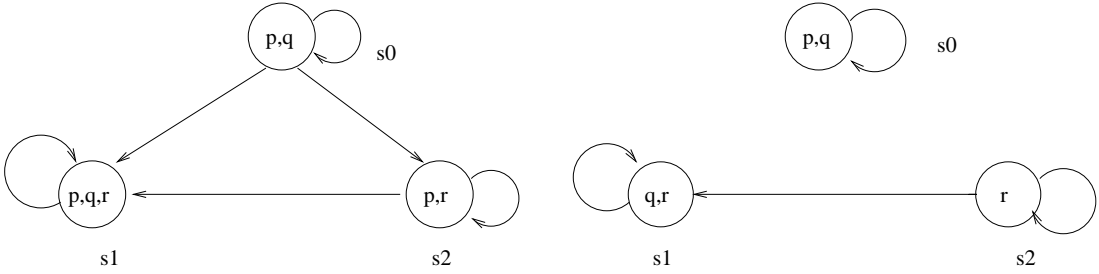}
\caption{Two possible results of updating $M$ with AG$p$.}
\label{ex1-1}
\end{center}
\end{figure}

The above example shows that in order to update a CTL model to satisfy a formula,
we may apply different kinds of operations to change the model. 
\comment{

Note that in Definition \ref{def:update}, we have not given what
the ordering $\leq_{M}$ is. Clearly, minimal change is not a 
domain-independent notion because with different contexts and requirements we
have different ways to define the required minimal change criterion and hence the
ordering $\leq_{M}$ as well. 

For our purpose,  
we would expect that the underlying minimal change principle is defined 
based on some operational process so that
a concrete algorithm for CTL model update can be implemented based
on this minimal change principle. 
}
From all possible operations applicable to a CTL model, we consider five basic ones 
where all changes on a CTL model can be achieved.

\comment{
Basically, as in traditional knowledge-base
update \cite{Winslett90}, we expect that a CTL model update obeys
an underlying minimal change principle. Furthermore, this minimal
change should be defined based on some operational process so that
a concrete algorithm for CTL model update can be implemented based
on this minimal change principle. To this end, we first propose
five primitive operations on the CTL model that provide a basis
for all complex CTL model updates.


The operations to update the CTL model can be decomposed into five
types, identified as PU1, PU2, PU3, and PU4. These primitive
updates are defined in their simplest forms as follows.

}

\vspace*{0.1in}
\noindent
{\bf PU1: Adding one relation element}\\ 
Given $M=(S,R,L)$, its updated model $M'=(S',R',L')$ is obtained
from $M$ by adding only one new relation element. That is, $S'=S$,
$L'=L$, and $R'=R\cup \{(s_{i},s_{j})\}$, where
$(s_{i},s_{j})\not\in R$ for two states $s_{i}, s_{j}\in S$.

\vspace*{0.1in}
\noindent
{\bf PU2: Removing one relation element}\\
Given $M=(S,R,L)$,  its updated model $M'=(S',R',L')$ is obtained
from $M$ by removing only one existing relation element. That is, $S'=S$,
$L'=L$, and $R'=R - \{(s_{i},s_{j})\}$, where $(s_{i},s_{j})
\in R$ for two states $s_{i}, s_{j}\in S$.

\vspace*{0.1in}
\noindent
{\bf PU3: Changing labeling function on one state}\\
Given $M=(S,R,L)$, its updated model $M'=(S',R',L')$ is obtained from
$M$ by changing labeling function on a particular state.  
That is, $S'=S$, $R'=R$, 
$\forall s\in (S-\{s^{*}\})$, $s^{*}\in S$, $L'(s)=L(s)$, and
$L'(s^{*})$ is a set of true variable assigned in state
$s^{*}$ where $L'(s^{*})\neq L(s^{*})$.

\vspace*{0.1in}
\noindent
{\bf PU4: Adding one state}\\
Given $M=(S,R,L)$, its updated model $M'=(S',R',L')$ is obtained
from $M$ by adding only one new state.
That is,  $S'=S\cup \{s^{*}\}$, $s^{*}\not\in S$,  $R'=R$, and 
$\forall s\in S$,
$L'(s)=L(s)$ and $L'(s^{*})$ is a set of true variables assigned in  $s^{*}$.

\vspace*{0.1in}
\noindent
{\bf PU5: Removing one isolated state}\\
Given $M=(S,R,L)$,  its updated model $M'=(S',R',L')$ is obtained
from $M$ by removing only one isolated state: 
$S'=S-\{s^{*}\}$, where $s^{*}\in S$ and $\forall s\in S$ such that $s\neq s^{*}$, 
neither $(s,s^{*})$ nor $(s^{*},s)$ is not in $R$,
$R'=R$, and $\forall s\in S'$, $L'(s)=L(s)$.

\vspace*{.1in}

\begin{figure}[tbhp]
\begin{center}
\epsfysize = 45 mm
\epsffile{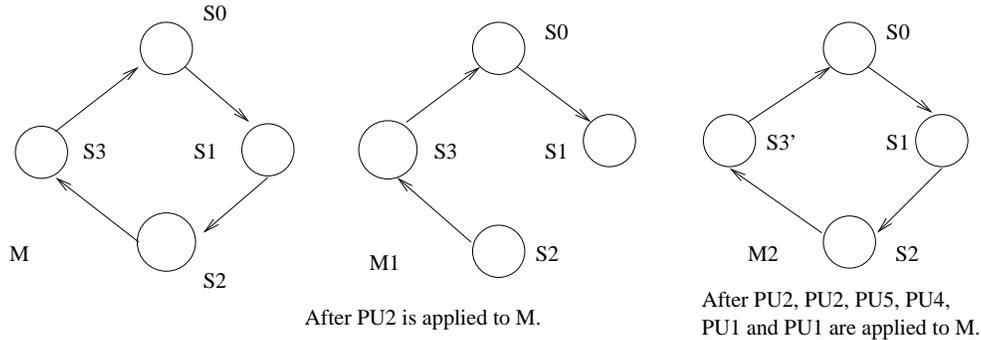}
\caption{Illustration of primitive updates.} 
\label{2.2}
\end{center}
\end{figure}

We call the above five operations {\em primitive} since they
express all kinds of changes to a CTL model. Figure \ref{2.2} illustrates
examples of applying some of these operations on a model.

In the above five operations, PU1, PU2, PU4 and PU5 represent the most basic
operations on a graph. Generally, using these four operations, we can 
perform any changes to a CTL model. For instance, if we want to substitute a state
in a CTL model, we do the following: (1) remove all relation elements associated to this state, 
(2) remove this isolated states, (3) add a state that we want to replace the original one,
and (4) add all relevant relation elements associated to this new state. 

Although these 
four operations are sufficient enough to represent all changes on a CTL model, they sometimes 
complicate the measure on the changes of CTL models. 
Consider the case of a state substitution. Given a CTL model $M$,
if one CTL model $M'$ has exactly the same graphical 
structure as $M$ except that $M'$ only has one particular state different
from $M$, then we tend to think
that $M'$ is obtained from $M$ with a single change of state replacement, 
instead of from a sequence of operations PU1, PU2, PU4 and PU5. 

This motivates us to have operation PU3. PU3 has an effect of state substitution, but it 
is fundamentally different from the combination of PU1, PU2, PU4 and PU5, because
PU3 does not change the state name and relation elements in the original model, it only
assigns a different set of propositional atoms to that state in the original
model. In this sense, the combination 
of PU1, PU2, PU4 and PU5 cannot replace operation PU3.
Using PU3 to represent state substitution significantly simplifies
our measure on the model difference as will be illustrated in Definition 
\ref{def:Close}. In the rest of the paper, we assume that 
all state substitutions in a CTL model will be achieved through PU3 so that
we have a unique way to measure the differences on CTL model changes in relation to
states substitutions.

We should also note that having operation PU3 as a way to substitute a state in a CTL model,
PU5 becomes unnecessary, because we actually do not need to 
remove an isolated state from a model. All we need is to remove relevant relation 
element(s) in the model, so that this state becomes unreachable from the initial state. 
Nevertheless, to remain our discussions to be coherent with 
all primitive operations described above, in the following 
definition on the CTL minimal change, we still consider the measure on 
changes caused by applying PU5 in a CTL model update.

\subsection{Defining Minimal Change}

Following traditional belief update
principle, in order to make a CTL model to satisfy some property, we would expect
that the given CTL model is changed as little as possible.
By using primitive update operations, a 
CTL Kripke model may be updated in different ways: 
adding or removing state transitions, adding new states, and changing the labeling
function for some state(s) in the model.
Therefore, we first need to have a method to measure 
the changes of CTL models, from which we can develop a minimal change criterion 
for CTL model update.

%
%
Given two CTL models $M=(S, R, L)$ and $M'=(S',R',L')$, for each
operation $PUi$ ($i=1, \cdots, 5$), 
${\mathit Diff}_{PUi}(M,M')$ denotes the differences between the two models
where $M'$ is an updated model from $M$, which makes
clear that several operations of type $PUi$ have occurred. Since PU1
and PU2 only change relation elements, we define 
${\mathit Diff}_{PU1}(M,M')=R'-R$ (adding relation elements only) and
${\mathit Diff}_{PU2}(M,M')=R-R'$ (removing relation elements only).
For operation PU3, since only labeling function is changed, the difference measure  
between $M$ and $M'$ for PU3 is defined as
${\mathit Diff}_{PU3}(M,M')= \{s\mid s\in S\cap S'$ and $L(s)\neq L'(s)\}$. 
For operations
PU4 and PU5, on the other hand, we define 
${\mathit Diff}_{PU4}(M,M')=S'-S$ (adding states) and
${\mathit Diff}_{PU5}(M,M')=S-S'$ (removing states).
%
Let ${\mathcal M}=(M,s)$ and ${\mathcal M'}=(M',s')$, for convenience, we 
also denote
 ${\mathit Diff}({\mathcal M},{\mathcal M'})$ $=$
$({\mathit Diff}_{PU1}(M,M'),{\mathit Diff}_{PU2}(M,M'),
{\mathit Diff}_{PU3}(M,M'), {\mathit Diff}_{PU4}(M,M'),
{\mathit Diff}_{PU5}(M,M'))$.

It is worth mentioning that given two CTL Kripke models
$M$ and $M'$, there is no ambiguity to compute
${\mathit Diff}_{PUi}(M,M')$ ($i=1,\cdots,5$), because each primitive operation will
only cause one type of changes (states, relation elements, or
labeling function) in the models no matter how many times it has been applied.
Now we can precisely define the ordering $\leq_{M}$ on CTL models.

\begin{definition}
\label{def:Close} ({\bf Closeness ordering}) Let 
$M$, $M_1$ and $M_2$ be three CTL Kripke models.
We say that $M_1$ is {\em  at least
as close to} $M$ {\em as} $M_2$, denoted as
$M_1\leq_{M}M_2$, if and only if for each set of PU1-PU5 operations that transform
$M$ to $M_2$, there exists a set of PU1-PU5 operations that transform
$M$ to $M_1$ such that 
the following conditions hold:
\begin{enumerate}
\item[(1)] for each $i$ ($i=1,\cdots,5$),
${\mathit Diff}_{PUi}(M,M_1)\subseteq {\mathit Diff}_{PUi}(M,M_2)$, and
\item[(2)] if ${\mathit Diff}_{PU3}(M,M_1)={\mathit Diff}_{PU3}(M,M_2)$, then for
each $s\in {\mathit Diff}_{PU3}(M,M_1)$, \\
$diff(L(s),L_1(s))\subseteq diff(L(s),L_2(s))$. 
\end{enumerate}
We denote $M_1<_M M_2$ if $M_1\leq_M M_2$ and $M_2\not\leq_M M_1$.
\end{definition}

Definition \ref{def:Close} presents a measure on the difference
between two models with respect to a given model.
Intuitively, we say that model $M_1$ is closer to $M$ relative to
model $M_2$, if (1) $M_1$ is obtained from $M$ by applying 
all primitive update operations that cause fewer
changes than those applied to obtain model $M_2$; and (2)
if the set of states in $M_1$ affected by applying PU3 
is the same as that in $M_2$, then we take a closer look at the difference 
on the set of propositional atoms associated with the relevant states.
Having the ordering specified in Definition \ref{def:Close}, we can define a 
CTL model update formally.

\begin{definition}
\label{def:admissible-update}
 ({\bf Admissible update}) Given a
CTL Kripke model $M=(S,R,L)$, ${\mathcal M}=(M,s_{0})$ where
$s_{0}\in S$, and a CTL formula $\phi$, a CTL Kripke model $Update({\mathcal
M},\phi)$ is called an {\em admissible} model (or admissible updated model)
if the following conditions hold: 
(1) $Update({\mathcal M},\phi)=(M',s_{0}')$, 
$(M',s_0')\models \phi$,
where $M'=(S',R',L')$ and $s_{0}'\in S'$; and, (2) there does not exist
another updated model $M''=(S'',R'',L'')$ and $s_{0}''\in S''$
such that $(M'',s_{0}'')\models \phi$ and $M''<_M M'$.  We use
${\mathit Poss}(Update({\mathcal M},\phi))$ to denote the set of all possible admissible
models of updating ${\mathcal M}$ to satisfy $\phi$.
\end{definition}

\begin{figure}[tbhp]
\begin{center}
\epsfysize = 40 mm
\epsffile{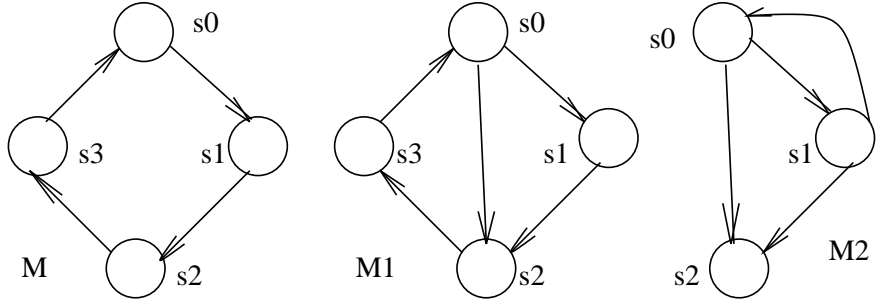}
\caption{Illustration of minimal change rules.}
\label{f:MimalChange-exp}
\end{center}
\end{figure}

\begin{example}
In Figure~\ref{f:MimalChange-exp}, model $M$ is updated in two
different ways. Model $M_1$ is the result of updating $M$ by
applying PU1. Model $M_2$ is another update of $M$ resulting by
applying PU1, PU2 and PU5. Then we have
${\mathit Diff}_{PU1}(M,M_1)=\{(s_0,s_2)\}$, and
${\mathit Diff}_{PU1}(M,M_2)=\{(s_1,s_0),(s_0,s_2)\}$, which results in 
${\mathit Diff}_{PU1}(M,M_1)\subset {\mathit Diff}_{PU1}(M,M_2)$. 
Also, it is easy to see that ${\mathit Diff}_{PU2}(M,M_1)=\emptyset$ and 
${\mathit Diff}_{PU2}(M,M_2)=\{(s_3,s_0), (s_2,s_3)\}$, so 
${\mathit Diff}_{PU2}(M,M_1)$ $\subset$ ${\mathit Diff}_{PU2}(M,M_2)$. 
Similarly, we can see that ${\mathit Diff}_{PU3}(M,M_1)
={\mathit Diff}_{PU3}(M,M_2)=\emptyset$, and  
${\mathit Diff}_{PU4}(M,M_1)= {\mathit Diff}_{PU4}(M,M_2)=\emptyset$. Finally, we have
${\mathit Diff}_{PU5}(M,M_1)=\emptyset$ and
${\mathit Diff}_{PU5}(M,M_2)=\{s_3\}$. 
 According to Definition \ref{def:Close}, we
have $M_1<_M M_2$.
\end{example}

We should note that in a CTL model update, if we can simply replace the initial state by
another existing state in the model to satisfy the formula, then this model actually has not
been changed, and it is the unique admissible model according to Definition 
\ref{def:admissible-update}. In this case, all other updates
will be ruled out by Definition \ref{def:admissible-update}.
For example, consider the CTL model $M$ described in Figure \ref{yan1}:
\begin{figure}[tbhp]
\begin{center}
\epsfysize = 50 mm
\epsffile{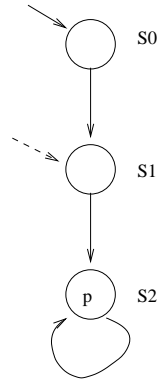}
\caption{A special model update scenario.}
\label{yan1}
\end{center}
\end{figure}
If we want to update $(M,s_0)$ with $\mbox{AX}p$, we can see that $(M,s_1)$ 
becomes the only admissible updated model according to our definition: 
we simply replace the initial state $s_0$ by $s_1$. Nevertheless,
we would expect that some other update may also be equally reasonable. For instance,
we may change the labeling function of $M$ to make $L'(s_1)=\{p\}$.
In both updates, we have changed something in $M$, but the change caused by the
first update is not represented in our minimal change definition.

We can overcome this difficulty by creating a {\em dummy state} $\sharp$ into a CTL Kripke 
model $M$, and for each initial state $s$ in $M$, we add relation
element $(\sharp,s)$ into $M$. In this way, a change of initial state from $s$ to 
$s'$ will imply a removal of relation element $(\sharp, s)$ and an addition of 
a new relation element $(\sharp,s')$. Such changes will be measured by our
minimal change definition. With this treatment, both updated models described above are
admissible. In the rest of the paper, without explicit declaration, we will assume 
that each CTL Kripke model contains a dummy state $\sharp$ and special
state transitions from $\sharp$ to all initial states. 

\section{Semantic Properties}
\label{sec:characterizations}

In this section, we first explore the relationship between our CTL model update
and traditional belief update, and then provide useful semantic characterizations on
some typical CTL model update cases.

\subsection{Relationship to Propositional Belief Update}

First we show the following result about ordering $\leq_{M}$ defined
in Definition \ref{def:Close}.

\begin{proposition}
$\leq_{M}$ is a partial ordering.
\label{ordering:prop}
\end{proposition}

\noindent
\begin{proof}
From Definition \ref{def:Close}, it is easy to see that $\leq_{M}$ is reflexive and
antisymmetric. Now we show that $\leq_{M}$ is also transitive.
Suppose $M_1\leq_{M} M_2$ and $M_2\leq_M M_3$. 
According to Definition \ref{def:Close}, we have 
$Diff_{PUi}(M,M_1)\subseteq Diff_{PUi}(M,M_2)$, and 
$Diff_{PUi}(M,M_2)\subseteq Diff_{PUi}(M,M_3)$ ($i=1, \cdots, 5$). Consequently,
we have $Diff_{PUi}(M,M_1)\subseteq Diff_{PUi}(M,M_3)$ ($i=1, \cdots, 5$).
So Condition 1 in  Definition \ref{def:Close} holds. 
Now consider Condition 2 in the definition. The only case we need to consider is that
$Diff_{PU3}(M,M_1)= Diff_{PU3}(M,M_2)$ and
$Diff_{PU3}(M,M_2)= Diff_{PU3}(M,M_3)$ (note that all other cases will directly 
imply $Diff_{PU3}(M,M_1)\subseteq Diff_{PU3}(M,M_3)$ and
$Diff_{PU3}(M,M_1)\neq Diff_{PU3}(M,M_3)$). 
In this case, it is obvious that for all $s\in Diff_{PU3}(M,M_1)$
$=$ $Diff_{PU3}(M,M_3)$,
$diff(L(s),L_1(s))\subseteq diff(L(s),L_3(s))$. So we have
$M_1\leq_{M} M_3$.
\end{proof}

It is also interesting to consider a special case of 
our CTL model update where the update formula is a 
classical propositional formula. The following proposition 
indicates that when only propositional formula is considered in CTL model update,
the admissible model can be obtained through the traditional model based
belief update approach \cite{Winslett88}.

\begin{proposition}
Let $M=(S,R,L)$ be a CTL model and $s_0\in S$. Suppose that $\phi$ is a 
satisfiable propositional formula and $(M,s_0)\not\models\phi$, 
then an admissible model of
updating $(M,s_0)$ to satisfy $\phi$ is $(M',s_0)$, 
where $M'=(S,R,L')$, for each $s\in (S-\{s_0\})$, $L'(s)=L(s)$,
$L'(s_0)\models\phi$, and there does not exist another $M''=(S,R,L'')$ such that
$L''(s_0)\models\phi$ and $diff(L(s_0),L''(s_0))\subset diff(L(s_0),L'(s_0))$.
\label{s:case}
\end{proposition}

\noindent
\begin{proof}
Since $\phi$ is a propositional formula, the update on $(M,s_0)$ to satisfy $\phi$ will 
not affect any relation elements and all other states except $s_0$. 
Since $L(s_0)\not\models\phi$, it is obvious that by applying PU3,
we can change the labeling function $L$ to $L'$
that assigns $s_0$ a new set of propositional atoms to satisfy $\phi$.
Then from Definition \ref{def:admissible-update}, we can see that the model specified in 
the proposition is indeed a minimally changed 
CTL model with respect to ordering $\leq_{M}$.
\end{proof}

We can see that the problem addressed by our
CTL model update is essentially different from the problem concerned in 
traditional propositional belief update. Nevertheless, the idea of model based
minimal change for CTL model update is closely related to  
belief update. Therefore,
it is worth investigating the relationship between 
our CTL model update and traditional propositional belief 
update postulates (U1) - (U8). 
In order to make such a comparison possible, 
we should lift the update operator
occurring in postulates (U1) - (U8) beyond the propositional logic case. 

For this purpose, we first introduce some notions.
Given a CTL formula $\phi$ and Kripke model $M=(S,R,L)$, let
$Init(S)\subseteq S$ be the set of all initial states in $M$.
$(M,s)$ is called a {\em model} of $\phi$ iff $(M,s)\models \phi$, where
$s\in Init(S)$. We use $Mod(\phi)$ to denote the set of all models of $\phi$.
Now we specify an update operator $\diamond_{c}$ to impose on CTL formulas
as follows: given two CTL formulas $\psi$ and $\phi$, we define that
$\psi\diamond_c\phi$ to be a CTL formula whose models are defined as:
\begin{quote}
$Mod(\psi\diamond_c\phi)=\bigcup_{(M,s)\in Mod(\psi)} 
{\mathit Poss}(Update((M,s),\phi))$.
\end{quote}

\begin{theorem}
Operator $\diamond_c$ satisfies all Katsuno and
Mendelzon update postulates (U1) - (U8).
\label{postulates}
\end{theorem}

\noindent
\begin{proof}
From Definitions 4 and 5, it is easy to verify that $\diamond_c$ satisfies 
(U1)-(U4). We prove that $\diamond_c$ satisfies (U5). 
To prove $(\psi\diamond_c\mu)\wedge\alpha\models \psi\diamond_c (\mu\wedge\alpha)$, 
it is sufficient to prove that for each model $(M,s)\in Mod(\psi)$, 
${\mathit Poss}(Update((M,s),\mu))\cap Mod(\alpha)\subseteq 
{\mathit Poss}(Update((M,s),\mu\wedge\alpha))$.
In particular, we need to show that for any 
$(M',s')\in {\mathit Poss}(Update((M,s),\mu))\cap Mod(\alpha)$, 
$(M',s')\in {\mathit Poss}(Update((M,s),\mu\wedge\alpha))$. 
Suppose $(M',s')\not\in {\mathit Poss}(Update((M,s),\mu\wedge\alpha))$. 
Then we have (1) $(M',s')\not\models\mu\wedge\alpha$; or
(2) there exists a different admissible
model $(M'',s'')\in Mod(\mu\wedge\alpha)$ such that
$M''<_M M'$. If it is case (1), then 
$(M',s')\not\in {\mathit Poss}(Update((M,s),\mu))\cap Mod(\alpha)$. So the result holds.
If it is case (2), it also implies that $(M'',s'')\models\mu$ and 
$M''<_M M'$. That means, $(M',s')\not\in {\mathit Poss}(Update((M,s),\mu))$. The result 
still holds.

Now we prove that $\diamond_c$ satisfies (U6). To prove this result, it is sufficient
to prove that for any $(M,s)\in Mod(\psi)$, if
${\mathit Poss}(Update((M,s),\mu_1)) \subseteq Mod(\mu_2)$ and
${\mathit Poss}(Update((M,s),\mu_2))$ $\subseteq$ $Mod(\mu_1)$, then
${\mathit Poss}(Update((M,s),\mu_1)) = {\mathit Poss}(Update((M,s),\mu_2))$.
We first prove
${\mathit Poss}(Update((M,s),\mu_1))$ $\subseteq$ ${\mathit Poss}(Update((M,s),\mu_2))$.
Let $(M',s')$ $\in$ ${\mathit Poss}(Update((M,s)$, $\mu_1))$. Then $(M',s')$ $\models$ $\mu_2$.
Suppose $(M',s')\not\in {\mathit Poss}(Update((M,s),\mu_2))$. Then there exists a different 
admissible model
$(M'',s'')\in {\mathit Poss}(Update((M,s),\mu_2))$ such that $M''<_M M'$. 
Also note that $(M'',s'')\models \mu_1$. 
This contradicts the fact
that $(M',s')\in {\mathit Poss}(Update((M,s),\mu_1))$.
So we have ${\mathit Poss}(Update((M,s),\mu_1))\subseteq {\mathit Poss}(Update((M,s),\mu_2))$.
Similarly, we can prove that 
${\mathit Poss}(Update((M,s),\mu_2)) \subseteq {\mathit Poss}(Update((M,s),\mu_1))$.

To prove that $\diamond_c$ satisfies (U7), it is sufficient to prove that 
${\mathit Poss}(Update((M,s), \mu_1)) \cap {\mathit Poss}(Update((M,s), \mu_1)) \subseteq 
{\mathit Poss}(Update((M,s), \mu_1\vee \mu_2))$, where
$(M,s)$ is the unique model of $T$ (note that $T$ is complete).
Let $(M',s')\in {\mathit Poss}(Update((M,s), \mu_1)) \cap {\mathit Poss}(Update((M,s)$, 
$\mu_1))$. 
Suppose $(M',s') \not\in {\mathit Poss}(Update((M,s), \mu_1\vee \mu_2))$. Then there exists
an admissible model $(M'',s'')\in {\mathit Poss}(Update((M,s), \mu_1\vee \mu_2))$ such that
$M''<_M M'$. Note that $(M'',s'')\models \mu_1\vee\mu_2$. 
If $(M'',s'')\models \mu_1$, then it implies that 
$(M',s')\not\in {\mathit Poss}(Update((M,s), \mu_1))$. 
If $(M'',s'')\models \mu_2$, then it implies 
$(M',s')\not\in {\mathit Poss}(Update((M,s), \mu_2))$. In both cases, we have 
$(M',s')\not\in {\mathit Poss}(Update((M,s), \mu_1))\cap {\mathit Poss}(Update((M,s), \mu_1))$.
This proves the result.

Finally, we show that $\diamond_c$ satisfies (U8). 
From Definition 5, we have that
$Mod((\psi_1\vee \psi_2)\diamond_c\mu)=\bigcup_{(M,s)\in Mod(\psi_1\vee \psi_2)}
{\mathit Poss}(Update((M,s),\mu))$ $=$
$\bigcup_{(M,s)\in Mod(\psi_1)}
{\mathit Poss}(Update((M,s),\mu))$ $\cup$
$\bigcup_{(M,s)\in Mod(\psi_2)}
{\mathit Poss}(Update((M,s),\mu))$ $=$ 
$Mod(\psi_1\diamond_c\mu)\cup Mod(\psi_2\diamond_c\mu)$.
This completes our proof.
\end{proof}

From Theorem 1, it is evident that 
Katsuno and Mendelzon's update postulates (U1) - (U8)
characterize a wide range of update formulations beyond the
propositional logic case, where model based
minimal change principle is employed. In this sense, we can view
that Katsuno and Mendelzon's update postulates (U1) - (U8) are essential
requirements for any model based update approaches.

\subsection{Characterizing Special CTL Model Updates}

From previous description, we observe that, for a
given CTL Kripke model $M$ and formula $\phi$, there may be many
admissible models satisfying $\phi$, where some are simpler
than others. In this section, we provide various results
that present possible
solutions to achieve admissible updates under certain conditions. In
general, in order to achieve admissible update results, we may have
to combine various primitive operations during an update process.
Nevertheless, as will be shown below, a single type primitive
operation will be enough to achieve an admissible updated model in
many situations. These characterizations also play an essential role
in simplifying CTL model update implementation.

Firstly, the following proposition simply shows that during a CTL update 
only reachable states will be taken into account in the sense that
unreachable state will never be removed or newly introduced.

\begin{proposition}
\label{pr-new}
Let $M=(S,R,L)$ be a CTL Kripke model, $s_0\in S$
an initial state of $M$, $\phi$ a satisfiable CTL formula
and $(M,s_0)\not\models \phi$. Suppose $(M',s_0')$ is an admissible
model after updating $(M,s_0)$ with $\phi$, where
$M'=(S',R',L')$. Then the following properties hold:
\begin{enumerate}
\item if $s$ is a state in $M$ (i.e. $s\in S$)
and is not reachable from $s_0$ (i.e. there does not
exist a path $\pi=[s_0,\cdots]$ in $M$ such that $s\in \pi$), then
$s$ must also be a state in $M'$ (i.e. $s\in S'$);

\item if $s'$ is a state in $M'$ and is not reachable from $s_0'$,
then $s'$ must also be a state in $M$.
\end{enumerate}
\end{proposition}

\noindent
\begin{proof}
We only give the proof of result 1 since the proof for result 2 is similar.
Suppose $s$ is not in $M'$. That is,
$s$ has been removed from $M$ during the generation of $(M',s_0')$.
From Definitions 4 and 5, we know that the only way to remove
$s$ from $M$ is to apply operation PU5 (and possibly other associated
operations such as PU2 - removing transition relations,
if $s$ is connected to other states).

Now we construct a new CTL Kripke model $M''$ in such a way
that $M''$ is exactly the same as
$M'$ except that $s$ is also in $M''$. That is,
$M''=(S'', R'',L'')$, where $S''=S'\cup\{s\}$,
$R''=R'$, for all $s^{*}\in S'$,
$L''(s^{*})=L'(s^{*})$, and $L''(s)=L(s)$.
Note that in $M''$, state $s$ is an isolated state, not connecting
to any other states.  Since $s$ is in $M$, from Definition 4 we can
see that $M''<_M M'$.
Now we will show that $(M'',s_0')\models\phi$.
We prove this by showing a bit more general result:
\begin{quote}
\underline{Result}: For any satisfiable CTL formula $\phi$ and any state $s^{*}\in S'$,
$(M'',s^{*})\models \phi$ iff $(M',s^{*})\models \phi$.
\end{quote}

This can be showed by induction on the structure of $\phi$.
(a) Suppose $\phi$ is a propositional formula. In this case,
$(M'',s^{*})\models \phi$ iff $L''(s^{*})\models \phi$. Since
$L''(s^{*})=L'(s^{*})$, and  $(M',s^{*})\models\phi$ iff
$L'(s^{*})\models \phi$, we have
$(M'',s^{*})\models\phi$ iff  $(M'',s^{*})\models \phi$.
(b) Assume that the result holds for formula $\phi$.
(c) We consider variours cases for formulas constructed from $\phi$.
(c.1) Suppose $\phi$ is of the form $\mbox{AG}\phi$.
$(M',s^{*})\models \mbox{AG}\phi$ iff for every path
from $s^{*}$ $\pi'=[s^{*},\cdots,]$, and for every state $s'\in \pi'$,
$(M',s')\models \phi$.  From the construction of $M''$, it is obvious
that every path from $s^{*}$ in $M'$ must be also a path in $M''$, and vice
versa. Also from the induction assumption, we have $(M',s')\models \phi$ iff
$(M'',s')\models \phi$. This follows that
$(M',s^{*})\models \mbox{AG}\phi$ iff $(M'',s^{*})\models \mbox{AG}\phi$.
Proofs for other cases such as $\mbox{AF}\phi$, $\mbox{EG}\phi$, etc.
are similar. 

Thus, we can find another model $M''$  such that
$(M'',s_0')\models \phi$ and $M''<_M M'$. This contradicts to the fact
that $(M',s_0')$ is an admissible model from the update of $(M,s_0)$
by $\phi$.
\end{proof}

\begin{theorem}
Let $M=(S,R,L)$ be a Kripke model and ${\mathcal
M}=(M,s_0)\not\models \mbox{\em EX}\phi$, where $s_0\in S$ and
$\phi$ is a propositional formula. Let 
${\mathcal M'}={\mathit Update}({\mathcal M}, \mbox{\em EX}\phi)$
be the model obtained from the update of ${\mathcal M}$ with $\mbox{\em EX}\phi$
through the following 1 or 2, then ${\mathcal M'}$ is an admissible model.
\begin{enumerate}
\item PU3 is applied to one $succ(s_0)$ to make 
$L'(succ(s_0))\models \phi$ and \\
${\mathit diff}(L(succ(s_0)),L'(succ(s_0)))$ minimal,
or, PU4 and PU1 are applied once successively to add a new state $s^{*}$ 
such that $L'(s^{*})\models\phi$ 
and a new relation element $(s_0,s^{*})$; 
\item if there exists some $s_i\in S$ such that
$L(s_i)\models \phi$ and $s_i\not = succ(s_{0})$, PU1 is applied once
to add a new relation element $(s_0,s_i)$.
\end{enumerate}
\label{th1}
\end{theorem}

\noindent
\begin{proof}
Consider case 1 first. After PU3 is applied to change the assignment on $succ(s_0)$, or
PU4 and PU1 are applied to add a new state $s^*$ and a relation element
$(s_0,s^{*})$, the new model $M'$ contains a $succ(s_0)$ such that
$L'(succ(s_0))\models\phi$. Thus, ${\cal M'}=(M',s_{0})\models
\mbox{EX}\phi$. If PU3 is applied once, then ${\mathit Diff}({\cal M}, {\cal
M'})=(\emptyset,\emptyset,\{succ(s_0)\}, \emptyset, \emptyset
)$; if PU4 and PU1 are applied once successively, ${\mathit Diff}({\mathcal M}$,
${\mathcal M'}) =(\{(s_0,s^{*})\},\emptyset,\emptyset,\{,s^*\},\emptyset )$. Thus,
updates by a single application of PU3 or applications of PU4 and PU1 once successively
are not compatible with each other. 
For PU3, if any other update is applied in
combination, ${\mathit Diff}({\cal M},{\cal M''})$ will either be not
compatible with ${\mathit Diff}({\cal M},{\cal M'})$ or contain 
${\mathit Diff}({\cal M},{\cal M'})$ (e.g., another PU3 together with its predecessor).
A similar situation occurs with the applications of PU4 and PU1. 
Thus, applying either PU3 once or
PU4 and PU1 once successively
represents a minimal change. For case 2, after PU1 is
applied to connect $s_0$ and $L(s_i)\models \phi$, the new model $M'$
has a successor which satisfies $\phi$. Thus, ${\cal
M'}=(M',s_0)\models \mbox{EX}\phi$. If PU1 is applied,
${\mathit Diff}({\mathcal M},{\mathcal M'}) =(\{(s_0,s_i)\}, \emptyset, \emptyset,
\emptyset, \emptyset )$. Note that this case remains a minimal change
of the relation element on the original model ${\mathcal M}$ and
is not compatible with case 1.
Hence, case 2 also represents a minimal change.
\end{proof}

Theorem \ref{th1} provides two cases where admissible CTL model
update results can be achieved for formula EX$\phi$. It is important
to note that here we restrict $\phi$ to be a propositional formula.
The first case says that we can either select one of the successor
states of $s_0$ and change its assignment minimally to satisfy
$\phi$ (i.e., apply PU3 once), or simply add a new state and a new relation element 
that satisfies $\phi$ as a successor of $s_0$ (i.e., apply PU4 and PU1 once successively). The
second case indicates that if some state $s_i$ in $S$ already
satisfies $\phi$, then it is enough to simply add a new relation element
$(s_0,s_i)$ to make it a successor of $s_0$. Clearly, both cases
will yield new CTL models that satisfy EX$\phi$.

\begin{theorem}
Let $M=(S,R,L)$ be a Kripke model and ${\mathcal M}=(M,s_0)\not
\models \mbox{\em AG}\phi$, where $s_0\in S$, $\phi$ is a
propositional formula and $s_0\models \phi$. 
Let ${\mathcal M'}={\mathit Update}({\mathcal M}, \mbox{\em AG}\phi)$ 
be a model obtained from the update of ${\mathcal M}$ with $\mbox{\em AG}\phi$ through 
the following way, then ${\mathcal M'}$ is an admissible model.
For each path starting from $s_0$: $\pi=[s_0,\cdots,s_i,\cdots]$:
\begin{enumerate}
\item if for all $s<s_i$ in $\pi$, $L(s)\models \phi$ but $L(s_i)\not\models\phi$,
PU2 is applied to remove relation element $(s_{i-1},s_i)$; or 
\item PU3 is applied to all states $s$ in $\pi$ not satisfying $\phi$ to change their
assignments such that $L'(s)\models\phi$ and ${\mathit diff}(L(s),L'(s))$ is minimal.
\end{enumerate}
\label{theoremAG}
\end{theorem}

\noindent
\begin{proof}
Case 1 is simply to cut path $\pi$ from the first state $s_i$
that does not satisfy $\phi$. Clearly, there is only one minimal way to cut
$\pi$: 
remove relation element $(s_{i-1},s)$ (i.e., apply PU2 once).
Case 2 is to minimally change the assignments for all states belonging to $\pi$
that do not satisfy $\phi$. Since the changes imposed by case 1 and case 2 are not
compatible with each other, both will generate admissible update results.
\end{proof}

In Theorem \ref{theoremAG}, case 1 considers a special form of the
path $\pi$ where the first $i$ states starting from $s_0$ already
satisfy formula $\phi$. Under this condition, we can simply cut off
the path to disconnect all other
states not satisfying $\phi$. Case 2 is straightforward: we
minimally modify the assignments of all states belonging to $\pi$ 
that do not satisfy formula $\phi$.


\begin{theorem}
\label{theorem:OrgEG} Let $M=(S,R,L)$ be a Kripke model, ${\cal
M}=(M,s_0)\not \models \mbox{\em{EG}}\phi$, where $s_0\in S$ and
$\phi$ is a propositional formula. 
Let ${\cal M'}={\mathit Update}({\cal M},\mbox{\em{EG}}\phi)$ be a model obtained
from the update of ${\cal M}$ with $\mbox{\em{EG}}\phi$ through the following way, then
${\cal M'}$ is an admissible model:
Select a path
$\pi=[s_0,s_1,\cdots,s_i,\cdots,s_j,\cdots]$ from $M$ which contains 
minimal number of different states not satisfying $\phi$\footnote{Note that
although a path may be infinite, it will only contain finite number of
different states.}, and then
\begin{enumerate}
\item if for all $s'\in \pi$ such that $L(s')\not\models \phi$, there
exist $s_i,s_j\in \pi$ satisfying $s_i<s'<s_j$ and $\forall s\leq
s_i$ or $\forall s\geq s_j$, $L(s)\models\phi$, then PU1 is applied
to add a relation element $(s_i,s_j)$, or PU4 and PU1 are applied to add a state
$s^{*}$ such that
$L'(s^{*})\models \phi$ and new relation elements $(s_i,s^{*})$ and
$(s^{*},s_j)$;

\item if $\exists s_i\in \pi$ such that $\forall s\leq s_i$,
$L(s)\models \phi$, and $\exists s_k\in \pi''$, where
$\pi''=[s_0,\cdots,s_k,\cdots]$ such that $\forall s\geq s_k$ and
$L(s)\models \phi$, then PU1 is applied to connect $s_i$ and $s_k$;

\item if $\exists s_i\in\pi$ ($i>1$) such that for all
$s'<s_i$, $L(s')\models\phi$, $L(s_i)\not\models\phi$, then,\\
 a. PU1 is applied to connect $s_{i-1}$ and $s'$ to form a new transition
 $(s_{i-1},s')$;\\
 b. if $s_i$ is the only successor of $s_{i-1}$, then
 PU2 is applied to remove relation element $(s_{i-1},s_i)$;

\item if $\exists$ $s'\in \pi$, such that $L(s')\not\models\phi$,
then PU3 is applied to change the assignments for all states $s'$ such that
$L'(s')\models \phi$ and ${\mathit diff}(L(s),L'(s'))$ is minimal.
\end{enumerate}
\end{theorem}

\noindent
\begin{proof}
In case 1, without loss of generality, we assume for the selected
path $\pi$, there exist states $s'$ that do not satisfy $\phi$, and
all other states in $\pi$ satisfy $\phi$. We also assume that such
$s'$ are in the {\em middle} of path $\pi$. Therefore, there are two
other states $s_i, s_j$ in $\pi$ such that $s_i<s'<s_j$. That is,
$\pi=[s_0,\cdots,s_{i-1},s_i,\cdots,s',\cdots,s_{j},s_{j+1},\cdots]$.
We first consider applying PU1. It is clear that by applying PU1 to
add a new relation element $(s_i,s_j)$, a new path is formed:
$\pi'=[s_0,\cdots,s_{i-1},s_i,s_j,s_{j+1},\cdots]$. Note that each
state in $\pi'$ is also in path $\pi$ and $s'\not\in\pi'$.
Accordingly, we know that $\mbox{EG}\phi$ holds in the new model $M'=(S,
R\cup\{(s_i,s_j)\},L)$ at state $s_0$. 
Consider ${\mathcal M}=(M,s_0)$ and ${\mathcal M'}=(M',s_0')$. 
Clearly, 
${\mathit Diff}({\mathcal M},{\mathcal M'})
=(\{(s_i,s_j)\},\emptyset, \emptyset, \emptyset, \emptyset)$,
which implies that $(M',s_0)$ must be a minimally changed model
with respect to $\leq_{M}$ that satisfies $\mbox{EG}\phi$. 

Now we consider applying PU4 and PU1. In this case, we will have a new model 
$M'=(S\cup\{s^{*}\}, R\cup \{(s_i,s^{*}), (s^{*},s_{j})\},L')$
where $L'$ is an extension of $L$ on new state $s^{*}$ that
satisfies $\phi$. We can see that
$\pi'=[s_0,\cdots,s_i,s^{*},s_{j},\cdots]$ is a path in $M'$ which
shares all states with path $\pi$ except the state $s^{*}$ in $\pi'$
and those states between $s_{i+1}$ and $s_{j-1}$ including $s'$ in
$\pi$. So we also have $(M',s_{0})\models \mbox{EG}\phi$.
Furthermore, we have ${\mathit Diff}({\mathcal M},{\mathcal M'})
=(\{(s_i,s^{*}), (s^{*},s_j)\}, 
\emptyset, \emptyset, \{s^{*}\},\emptyset)$.
Obviously, $(M',s_0)$ is a minimally changed model with respect to
$\leq_{M}$ that satisfies EG$\phi$.

It is worth mentioning that in case 1, the model obtained by  only applying PU1
is not comparable to the model obtained by applying PU4 and PU1, because 
no set inclusion relation holds for the changes on relation elements 
caused by these two different ways.

In case 2, consider two different paths
$\pi=[s_0,\cdots,s_i,\cdots]$ and $\pi'=[s_0,\cdots,s_k,\cdots]$
such that all states before state $s_i$ in path $\pi$ satisfy
$\phi$, and all states after state $s_k$ in path $\pi'$ satisfy
$\phi$, then PU1 is applied to form a new transition $(s_i,s_k)$.
This transition therefore connects all states from $s_0$ to $s_i$ in
path $\pi$  and all states after $s_k$ in path $\pi'$. Hence
all states in the new path $[s_0,\cdots,s_i,s_k\cdots]$ satisfy
$\phi$. Thus, $\cal M$$'\models \mbox{EG}\phi$. Such change is also
minimal, because after PU1 is applied, 
${\mathit Diff}({\mathcal M},{\mathcal M'})
=(\{(s_i,s_k)\},\emptyset,\emptyset,\emptyset,\emptyset)$ is
minimum and $(M',s_0)$ is a minimally changed model with respect to
$\leq_{M}$ that satisfies EG$\phi$.

In case 3, there are two situations. (a) If PU1 is applied to form a
new transition $(s_{i-1},s')$, then a new path containing
$[s_0,\cdots,s',\cdots,s_{i-1},s',\cdots,s_{i-1},s',\cdots]$
consists of Strongly Connected Components
where all states satisfy $\phi$, and \\
${\mathit Diff}({\mathcal M},{\mathcal M'})$
$=$ $(\{(s_{i-1},s')\},\emptyset, \emptyset,\emptyset,\emptyset)$
is minimum. Thus, $(M',s_0)$ is a minimally changed model with
respect to $\leq_{M}$ that satisfies EG$\phi$.

(b) If PU2 is applied, then, a new path $\pi'$ containing
$[s_0,\cdots,s',\cdots,s_{i-1}]$ is derived where all states satisfy
$\phi$ and $Diff(M,M')=(\emptyset,
\{(s_{i-1},s_i)\},\emptyset,\emptyset,\emptyset)$ is minimal.
Obviously, $(M',s_0)$ is a minimally changed model with respect to
$\leq_{M}$ that satisfies $\mbox{EG}\phi$. 
\comment{
On the other hand, if PU5
is applied, then, the same new path $\pi'$ is formed and ${\mathit
Diff}($$\cal M$,$\cal
M$$')=(\emptyset,\emptyset,\emptyset,\emptyset,\{s_i\})$ is minimum.
Obviously, $\cal M$$'$ is a minimally changed model with respect to
$\leq_{M}$ that satisfies $\mbox{EG}\phi$.
}

In case 4, suppose that there are $n$ states on the selected path
$\pi$ that do not satisfy $\phi$. After PU3 is applied to all these
states, ${\mathit Diff}({\mathcal M},{\mathcal M'})=(\emptyset,\emptyset,\{
s'_1,s'_2,\cdots,s'_n\},\emptyset,\emptyset)$, 
where for each $s'\in \{s_1',\cdots,s_n'\}$, $diff(L(s'),L'(s'))$ is minimal.
${\mathit Diff}({\mathcal M}, {\mathcal M'})$ in this case is not compatible with
those in cases 1, 2 and 3. Thus, $(M',s_0)$ is a minimally changed
model with respect to $\leq_{M}$ that satisfies $\mbox{EG}\phi$.
\end{proof}

Theorem \ref{theorem:OrgEG} characterizes four typical situations
for the update with formula EG$\phi$ where
$\phi$ is a propositional formula. Basically, this theorem says
that in order to make formula EG$\phi$ true, we first select a
path, then we can either make a new path based on this path so
that all states in the new path satisfy $\phi$ (i.e., case 1, case
2 and case 3(a)), or trim the path from the state where all
previous states satisfy $\phi$ (i.e., case 3(b)), if the previous
state has only this state as its successor; or simply change the 
assignments for all states not satisfying $\phi$ in the path
(i.e., case 4). Our proof shows that 
models obtained from these operations are admissible.

It is possible to provide further semantic characterizations for
updates with other special CTL formulas such as EF$\phi$, AX$\phi$,
and E$[\phi\mbox{U} \psi]$. In fact, in our prototype implementation,
such characterizations have been used to simplify the update process
whenever certain conditions hold. 

We should also indicate that all characterization theorems presented
in this section only provide sufficient conditions to compute admissible models. There
are other admissible models which will not be captured by these
theorems.


\section{Computational Properties}
\label{sec:ComputationalProp}

In this section, we study 
computational properties for our CTL model
update approach in some detail. We will first present a general complexity
result, and then we identify a useful subclass of CTL model updates which can always
be achieved in polynomial time.

\subsection{The General Complexity Result}

\begin{theorem}
Given two CTL Kripke models $M=(S,R,L)$ and $M'=(S',R',L')$, where
$s_0\in S$ and $s_0'\in S'$,  and a CTL formula $\phi$, it is
co-NP-complete to decide whether $(M',s_0')$ is an admissible
model of the update of $(M,s_{0})$ to satisfy $\phi$. The
hardness holds even if $\phi$ is of the form $\mbox{\em EX}\psi$
where $\psi$ is a propositional formula. \label{thNP}
\end{theorem}

\noindent
\begin{proof}
Membership proof: Firstly, we know from Clarke et al. \citeyear{Clarke&etal99} that
checking whether $(M',s_0')$ satisfies $\phi$ or not can be
performed in time ${\mathcal O}(|\phi|\cdot (|S|+|R|))$. In order
to check whether $(M',s_0')$ is an admissible update result, we
need to check whether $M'$ is a minimally updated model with
respect to ordering $\leq_{M}$. For this purpose, we consider the
complement of the problem by checking whether $M'$ is {\em not} a
minimally updated model. Therefore, we do two things: (1) guess
another updated model of $M$:  $M''=(S'',R'',L'')$ satisfying $\phi$ for some $s''\in
S''$; and, (2) test whether $M''<_{M} M'$. Step (1) can be done in
polynomial time. To check $M''<_{M} M'$, we first compute
$diff(S,S')$, $diff(S,S'')$, $diff(R,R')$ and $diff(R,R'')$. All
these can be computed in polynomial time. Then, according to these
sets, we identify $Diff_{PUi}(M,M')$ and $Diff_{PUi}(M,M'')$
($i=1,\cdots,5$) in terms of PU1 to PU5. Again, these steps can
also be completed in polynomial time. Finally, by checking
$Diff_{PUi}(M,M'')\subseteq Diff_{PUi}(M,M')$ ($i=1,\cdots,5$), and
$diff(L(s),L'(s))\subseteq diff(L(s),L''(s))$ for 
all $s\in Diff_{PU3}(M,M'')$ (if $Diff_{PU3}(M,M'')=Diff_{PU3}(M,M')$), we can
decide whether $M''<_{M} M'$. Thus, both steps (1) and (2) can be
achieved in polynomial time with a non-deterministic Turing
machine.

Hardness proof: It is well known that the validity problem for a
propositional formula is co-NP-complete. Given a propositional
formula $\phi$, we construct a transformation from the problem of
deciding $\phi$'s validity to a CTL model update in polynomial time.
Let $X$ be the set of all variables occurring in $\phi$, and $a,b$
two new variables do not occur in $X$. We denote $\neg
X=\bigwedge_{x_i\in X}\neg x_i$. Then, we specify a CTL Kripke model
based on the variable set $X\cup \{a,b\}$: $M=(\{s_0,s_1\},
\{(s_0,s_1),(s_1,s_1)\}, L)$, where $L(s_0)=\emptyset$ (i.e., all
variables are assigned false), $L(s_1)=X$ (i.e., variables in $X$
are assigned true, while $a,b$ are assigned false). Now we define a
new formula $\mu=\mbox{EX}(((\phi\supset a)\wedge (\neg X \wedge
b))\vee (\neg \phi \wedge a))$. Clearly, formula $((\phi\supset
a)\wedge (\neg X \wedge b))\vee (\neg \phi \wedge a)$ is satisfiable
and $s_1\not\models ((\phi\supset a)\wedge (\neg X \wedge b))\vee
(\neg \phi \wedge a)$. So $(M,s_0)\not\models \mu$.
Consider the update $Update((M,s_0),\mu)$. We define
$M'=(\{s_0,s_1\},\{(s_0,s_1),(s_1,s_1)\}, L')$, where
$L'(s_0)=L(s_0)$ and $L'(s_1)=\{a,b\}$. Next, we will show that
$\phi$ is valid iff $(M',s_0)$ is an admissible update result from $Update((M,s_0),\mu)$. \\
\underline{Case 1}: We show that if $\phi$ is valid, then
$(M',s_0)$ is an admissible update result from
$Update((M,s_0),\mu)$. Since $\phi$ is valid, we have $\neg X\models
\phi$. Thus, $L'(s_1)\models (\phi\supset a)\wedge (\neg X \supset
b))$. This leads to $(M',s_{0})\models \mu$. Also note that $M'$ is
obtained by applying PU3 to change $L(s_1)$ to $L'(s_1)$.
$diff(L(s_1),L'(s_1))=X\cup \{a,b\}$, which presents a minimal change from
$L(s_1)$ in order to satisfy $(\phi\supset a)\wedge (\neg X \wedge b)$. \\
\underline{Case 2}: Suppose that $\phi$ is not valid. Then,
$X_1\subseteq X$ exists such that $X_1\models\neg \phi$. We
construct
$M''=(\{s_0,s_1\},\{(s_0,s_1),(s_1,s_1)\},L'')$, where
$L''(s_0)=L(s_0)$ and $L''(s_1)=X_1\cup\{a\}$. It can be seen
that $L''(s_1)\models (\neg \phi\wedge a)$, hence $(M'',s_0)\models
\mu$. Now we show that $(M',s_0)\models\mu$ implies $M''<_{M}
M'$. Suppose $(M',s_0)\models\mu$. Clearly, both $M'$ and $M''$
are each obtained from $M$ by applying PU3 once to change the assignment on $s_1$. However,
we have $diff(L(s_1),L''(s_1))=(X-X_1)\cup\{a\}\subset
X\cup\{a,b\}=diff(L(s),L'(s_1))$.
Thus, we conclude that $(M',s_0)$ is not an admissible updated model.
\end{proof}

Theorem \ref{thNP} implies that it is probably not feasible to develop a polynomial time
algorithm to implement our CTL model update. Indeed, our algorithm described in
the next section, generally
runs in exponential time. 

\subsection{A Tractable Subclass of CTL Model Updates}

In the light of the complexity result of Theorem \ref{thNP}, we expect to identify 
some useful cases of CTL model updates which can be performed efficiently. 
First, we have the following observation.

\vspace*{.1in}
\noindent
{\bf Observation}: 
Let $M=(S,R,L)$ be a CTL Kripke model, $\phi$ a CTL formula and
$(M,s_0)\not\models\phi$ where $s_0\in S$.
If an admissible model
$Update((M,s_0),\phi)$ is 
obtained by only applying operations PU1 and PU2 to $M$,
then this result can be computed in polynomial time.

\vspace*{.1in}
Intuitively, if an admissible
updated model can be obtained by only using PU1 and PU2,
then it implies that we only need to
at most visit all states and relation elements in
$M$, and each operation involving PU1 or PU2
can be completed by just adding or removing relation elements, 
which obviously can be done in linear time. 

This observation tells us that under certain conditions,
operations PU1 and PU2 may be efficiently applied to compute
an admissible model. This is quite obvious because both PU3 and PU4 are involved
in finding models for some propositional formulas, while applying PU3 usually needs to further
find the minimal change on the assignment on the state, both of these operations
may cost exponential time in the size of 
input updating formula $\phi$. 
However, the above observation 
does not tell us what kinds of CTL model updates can really be achieved
in polynomial time. In the following, we will provide a sufficient
condition for a class of CTL model updates
which can always be solved in polynomial time.

We first specify a subclass of CTL formulas {\bf AEClass}:
(1) formulas AX$\phi$, AG$\phi$, AF$\phi$, A$[\phi_1\mbox{U} \phi_2]$,
EX$\phi$, EG$\phi$, EF$\phi$ and E$[\phi_1\mbox{U} \phi_2]$ are
in {\bf AEClass}, where $\phi$, $\phi_1$ and $\phi_2$ are propositional formulas;
(2) if $\psi_1$ and $\psi_2$ are in {\bf AEClass}, then 
$\psi_1\wedge\psi_2$ and $\psi_1\vee\psi_2$ are in {\bf AEClass}; 
(3) no formulas other than those specified in (1) and (2) are in {\bf AEClass}.
We also call formulas of the forms specified in (1) are {\em atomic} {\bf AEClass} 
formulas.

Note that {\bf AEClass} is a class of CTL formulas without nested temporal operators.
Although this is somewhat restricted, as we will show next, updates with this kind of CTL 
formulas may be much simpler than other cases.
Now we define valid states and paths for {\bf AEClass} formulas
with respect to a given model.

\begin{definition}
({\bf Valid state and path for AEClass})
Let $M=(S,R,L)$ be a CTL Kripke model, $\psi\in\mbox{\bf AEClass}$, and
$(M,s_0)\not\models\psi$, where $s_0\in S$.
We define $\psi$'s {\em valid state} or {\em valid path} 
in $(M,s_0)$ as follows.
\begin{enumerate}
\item If $\psi$ is of the form $\mbox{\em AX}\phi$, then state $s\in S$ is a {\em valid state}
of $\psi$ in $(M,s_0)$ if $(s_0,s)\in R$ and $L(s)\models\phi$;
\item If $\psi$ is of the form (a) $\mbox{\em AG}\phi$, 
(b) $\mbox{\em AF}\phi$ or (c) $\mbox{\em A}[\phi_1\mbox{\em U} \phi_2]$, 
then a path $\pi=[s_0,\cdots]$ is a {\em valid path} of $\psi$ 
in $(M,s_0)$ if $\forall s\in\pi$, $L(s)\models \phi$ (case (a));
$\exists s\in\pi$ and $s>s_0$, $L(s)\models\phi$ (case (b)); or $\exists s\in\pi$,
$s\models\phi_2$ and $\forall s'<s$ $L(s')\models\phi_1$ (case (c)) respectively;
\item If $\psi$ is of the form $\mbox{\em EX}\phi$, then state $s\in S$ is a {\em valid state}
of $\psi$ in $(M,s_0)$ if $L(s)\models\phi$;
\item If $\psi$ is of the form (a) $\mbox{\em EG}\phi$, 
(b) $\mbox{\em EF}\phi$ or (c) $\mbox{\em E}[\phi_1\mbox{\em U} \phi_2]$,
then a path $\pi=[s_0',\cdots]$ ($s_0'\neq s_0$) is a {\em valid path} of $\psi$ in
$(M,s_0)$ if $\forall s\in\pi$, $L(s)\models \phi$ and $L(s_0)\models\phi$ (case (a));
$\exists s\in\pi$ and $s>s_0'$, $L(s)\models\phi$ (case (b)); or 
$L(s_0)\models \phi_1$ and $\exists s\in\pi$,
$L(s)\models\phi_2$ and $\forall s'<s$ $L(s')\models\phi_1$ (case (c)) respectively.
\end{enumerate}
For an arbitrary $\psi\in \mbox{\bf AEClass}$, we say that $\psi$
has a {\em valid witness} in $(M,s_0)$ if every atomic {\bf AEClass} formula occurring
in $\psi$ has a valid state or path in $(M,s_0)$.
\label{validAE}
\end{definition}

Intuitively, for formulas $\mbox{AX}\phi$,
$\mbox{AG}\phi$, $\mbox{AF}\phi$ and
$\mbox{A}[\phi_1\mbox{U} \phi_2]$, a valid state or path in a CTL model
represents a local structure that partially satisfies the underlying formula.
For formulas $\mbox{EX}\phi$,
$\mbox{EG}\phi$, $\mbox{EF}\phi$ and
$\mbox{E}[\phi_1\mbox{U} \phi_2]$, on the other hand, 
a valid state or path also represents a local structure which {\em will}  satisfy
the underlying formula if a relation element is added to connect this local structure
and the initial state. 

\begin{example}
Consider the CTL Kripke model $M$ in 
Figure \ref{fig-y1} and formula $\mbox{EX} (p\wedge q)$. Clearly, 
$(M,s_0)\not\models \mbox{EX} (p\wedge q)$. Since $p,q\in L(s_3)$, $s_3$ is a
valid state of $\mbox{EX} (p\wedge q)$. Then we can simply add one 
relation element
$(s_0,s_3)$ into $M$ to form a new model $M'$ so that $(M',s_0)\models \mbox{EX} (p\wedge q)$.
Obviously, $(M',s_0)$ is an admissible updated model.

\begin{figure}[tbhp]
\begin{center}
\epsfysize = 50 mm
\epsffile{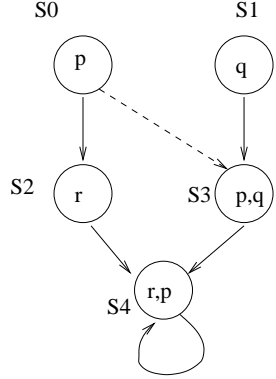}
\caption{A simple CTL model update.} 
\label{fig-y1}
\end{center}
\end{figure}
\end{example}

From the above example, we observe that if we update a CTL model with an {\bf AEClass} 
formula and this formula has a valid witness in the model,
then it is possible to compute an admissible model by 
only adding or removing relation elements (i.e. operations PU1 and PU2).
The following results confirm that a CTL model update with 
an {\bf AEClass} formula may be achieved in 
polynomial time if the formula has a valid witness in the model.

\begin{theorem}
Let $M=(S,R,L)$ be a CTL Kripke model, $\psi\in\mbox{\bf AEClass}$,
and $(M,s_0)\not\models\psi$. Deciding whether $\psi$ has a valid witness in $(M,s_0)$
can be solved in polynomial time. Furthermore, if $\psi$ has a valid witness in $(M,s_0)$,
then all valid states and paths of atomic {\bf AEClass} 
formulas occurring in $\psi$ can be computed from $(M,s_0)$ 
in time ${\mathcal O}(|\psi|\cdot (|S|+|R|)^{2})$.
\label{valid}
\end{theorem}

\noindent
\begin{proof}
To prove this theorem, we show that by using 
CTL model checking algorithm SAT \cite{HuthandRyan2000}, which takes
a CTL Kripke model and
an {\bf AEClass} formula as inputs, we can generate all valid states and paths of 
atomic {\bf AEClass} formulas occurring in $\psi$ (if any). We know that
the complexity of algorithm SAT is ${\mathcal O}(|\psi|\cdot (|S|+|R|))$.
We consider each case of atomic {\bf AEClass} formulas.

$\psi$ is $\mbox{AX}\phi$. We use SAT to check whether
$(M,s_0)\models \mbox{EX}\phi$. If $(M,s_0)\not\models \mbox{EX}\phi$, then
$\mbox{AE}\phi$ does not have a valid state in $(M,s_0)$. Otherwise, SAT will
return a state $s$ such that $L(s)\models \phi$ and $(s_0,s)\in R$. Then remove 
relation  element $(s_0,s)$ from $M$, and continue checking formula $\mbox{EX}\phi$ in the
model. By the end of this process, we obtain all valid states in 
$(M,s_0)$ for formula $\mbox{AX}\phi$. Altogether, there are at most $|S|$ SAT calls.

$\psi$ is $\mbox{AG}\phi$. We use SAT to check whether
$(M,s_0)\models \mbox{EG}\phi$. If $(M,s_0)\models \mbox{EG}\phi$, then we can obtain
a path in $M$ from SAT $\pi=[s_0,s_1,\cdots]$ such that $\forall s\in \pi$, $L(s)\models \phi$.
Clearly, such $\pi$ is a valid path of $\mbox{AG}\phi$.
Now if there does not exist a state $s^{*}$ such that $s^{*}\not\in \pi$ and $(s,s{*})\in R$ 
for some $s\in\pi$, 
i.e. state $s$ connects to state $s^{*}$ leading to a different path, then the process
stops, and $\pi$ is the only valid path for $\mbox{AG}\phi$. Otherwise,
Then we remove one relation element $(s,s')$ from $\pi$ (i.e. $s,s'\in \pi$) such that 
for all states $s''\in \pi$ where $s'<s''$, there is no relation element $(s'',s^{*})$ leading to
a different path (i.e. $s^{*}\not\in \pi$). In this way, we actually disable path $\pi$ to 
satisfy formula $\mbox{EG}\phi$ without affecting other paths. 
Then we continue checking formula 
$\mbox{EG}\phi$ in the newly obtained model. By the end of this process, we will obtain all
paths that make $\mbox{EG}\phi$ true, and these paths are all valid paths for
$\mbox{AG}\phi$. Since for each generated valid path, we need to remove one
relation element from this path before we generate the next valid path, there are at most 
$|R|$ such valid paths to be generated. So all together, there are at most $|R|$ SAT
calls to find all valid paths of $\mbox{AG}\phi$.

In the cases of  
$\mbox{AF}\phi$ and $\mbox{A}[\phi\mbox{U} \phi_2]$, all valid paths for these formulas can 
be generated in a similar way as described above for formula $\mbox{AG}\phi$.
The only different point is that for the case of $\mbox{A}[\phi\mbox{U} \phi_2]$, 
once a valid path $\pi$ has been generated, we need to find the last state 
$s\in\pi$ before $\phi_2$ becomes true, such that $s$ connects to a state $s^{*}\not\in\pi$
leading to a different path, then we disable $\pi$ by removing relation element
$(s,succ(s))$ from $\pi$. Then we continue the procedure to generate the next valid path
for $\mbox{A}[\phi\mbox{U}\phi_2]$.
If no such $s$ exists in $\pi$, then the process stops.

$\psi$ is $\mbox{EX}\phi$. In this case, each valid state $s$ can be found by
checking whether $L(s)\models \phi$. 
At most we need to visit $|S|$ states for this checking.

$\psi$ is $\mbox{EG}\phi$. Similarly, we can find a valid path  by selecting
a state $s\in S$ ($s\neq s_0$),
such that $(M,s)\models \mbox{EG}\phi$. At most, we need to visit $|S|$ states, and have
$|S|$  SAT calls 
to check $(M,s)\models \mbox{EG}\phi$. 

Finally, valid paths for $\mbox{EF}\phi$ and $\mbox{E}[\phi_1\cup\phi_2]$ can be 
found in a similar way.
\end{proof}

\begin{theorem}
Let $M=(S,R,L)$ be a CTL Kripke model, $\psi\in\mbox{\bf AEClass}$, 
and $(M,s_0)\not\models\psi$. 
An admissible model $Update((M,s_0),\psi)$ can be computed in 
polynomial time if $\psi$ has a valid witness in $(M,s_0)$.
\label{tract}
\end{theorem}

\noindent
\begin{proof}
From the proof of Theorem \ref{valid}, we can obtain all valid states and paths for 
all atomic {\bf AEClass} formulas in $\psi$ in time ${\mathcal O}(|\psi|\cdot (|S|+|R|)^{2})$.
Now we consider each case of atomic {\bf AEClass} formulas $\psi$, while the cases of
conjunctive and disjunctive {\bf AEClass} formulas are easy to justify.

$\psi$ is $\mbox{AX}\phi$. Let $S^{*}=\{s_1,\cdots,s_k\}$ be all valid 
states for $\mbox{AX}\phi$. Then we remove all 
relation elements $(s_0,s)$ where $s\not\in S^{*}$. 
In this way, we obtain a new model 
$M'=(S,R',L)$, where $R'=R-\{(s_0,s)\mid s\not\in S^{*}\}$. 
Obviously, we have $(M',s_0)\models \mbox{AX}\phi$. It is also easy to see that the
change between $M$ and $M'$ is minimal in order to 
satisfy $\mbox{AX}\phi$. So $(M',s_0)$ is also an admissible model.

$\psi$ is $\mbox{AG}\phi$. Let $S^{*}$ be the set of all states 
that are in some valid paths of $\mbox{AG}\phi$. For each state $s'\in S$ such that
$L(s')\not\models \phi$, we check whether $s'$ is reachable from $s_0$. If it is reachable,
then we remove a relation element $(s_1,s_2)$ from $M$ so that 
$s'$ becomes unreachable from $s_0$ and $(s_1,s_2)$ is not a relation element in a valid path
of $\mbox{AG}\phi$. 
Clearly, model $(M',s_0)$ will then satisfy $\mbox{AG}\phi$. 
Also, checking whether a state is reachable from $s_0$ can be done in polynomial time
by computing a spanning tree of $M$ rooted at $s_0$ \cite{tree02}.

$\psi$ is $\mbox{AF}\phi$. 
In this case, we need to cut all those paths starting from $s_0$ that are not valid 
paths for $\mbox{AF}\phi$ in $(M,s_0)$.
For doing this, it is sufficient to disconnect all states that are reachable from 
$s_0$ but not occur in any of $\mbox{AF}\phi$'s valid paths in $(M,s_0)$.
Let $S^{*}$ be the set of all these states, and 
$R^{*}$ be the set of all relation elements that are directly connected to these states,
i.e. $(s_1,s_2)\in R^{*}$ iff $s_1\in S^{*}$ or $s_2\in S^{*}$. Then we remove a minimal 
subset of $R^{*}$ from $M$ such that removing them will disconnect all states
in $S^{*}$ from $s_0$. The set $S^{*}$ can be identified in polynomial time 
by computing a spanning tree of $M$ rooted at $s_0$, and the minimal subset of $R^{*}$
that disconnects all states $S^{*}$ from $s_0$ can be found in time
${\mathcal O}(|R^{*}|^{2})$. So the entire process can be completed in polynomial time.

The case of $\mbox{A}[\phi_1\mbox{U} \phi_2]$ can be handled in a similar way  as described above 
for $\mbox{AF}\phi$.

Now we consider that $\psi$ is 
$\mbox{EX}\phi$. In this case, we only need to select one valid state $s$ for $\mbox{EX}\phi$, 
and add relation element $(s_0,s)$ into $M$. Then the model $(M',s_0)$ satisfies $\mbox{EX}\phi$.
For the case of $\mbox{EG}\phi$, we also select a valid path $\pi=[s,\cdots]$ for $\mbox{EG}\phi$,
and then add a relation element 
$(s_0,s)$, so we have $(M',s_0)\models \mbox{EG}\phi$. The other two 
cases of $\mbox{EF}\phi$ and $\mbox{E}[\phi_1\mbox{U}\phi_2]$ can be handled in a similar way.
\end{proof}

We should emphasize that although the above results characterize a useful subclass
of CTL model update scenarios in which some admissible updated models can be computed
through simple operations of adding or removing relation elements, 
it does not mean that all 
such admissible models represent intuitive modifications from a practical viewpoint.
Sometimes, for the same update problem,
using other operations such as PU3 and PU4 are probably more preferred in order
to generate a sensible system modification. This will be illustrated in Section 7.


\section{CTL Model Update Algorithm}
\label{sec:Algorithms}

\label{sec:mainfunction}

We have implemented a prototype for the CTL model update.
In this implementation, the CTL model update algorithm is designed
in line with the CTL model checking algorithm used in SAT
\cite{HuthandRyan2000}, where an updated formula is parsed
according to its structure and recursive calls to appropriate
functions are used. This recursive call usage allows the checked
property $\phi$ to range from nested modalities to atomic
propositional formulas. 
In this section, we will focus our discussions
on  the key ideas of handling CTL model update and provide high level 
pseudo code for major functions in the algorithm. 

\subsection{Main Functions}

\noindent
{\bf Handling propositional formulas}

\vspace*{.1in}
Since the satisfaction of a propositional formula does not involve any relation elements
in a CTL Kripke model, we implement the update with a propositional formula directly
through operation PU3 with a minimal change on the labeling function of the truth
assignment on the relevant state. This procedure is outlined as follows.

\vspace*{.1in}
\noindent
$\ast$ function $\mbox{Update}_{prop}((M,s_0), \phi)$ $\ast$\\
input: $(M,s_0)$ and $\phi$, where $M=(S,R,L)$ and $s_0\in S$;\\
output: $(M',s_0')$, where $M'=(S',R',L')$, $s_0'\in S'$ and $L'(s_0')\models\phi$;\\
{\tt 01} \hspace*{.1in} {\bf begin}\\
{\tt 02} \hspace*{.2in} apply PU3  to change labeling function $L$ on
state $s_0$ to form a new model $M'=(S',R',L')$:\\
{\tt 03} \hspace*{.4in} $S'=S$; $R'=R$; $\forall s\in S$ that $s\neq s_0$, $L'(s)=L(s)$;\\
{\tt 04} \hspace*{.4in} $L'(s_0)$ is defined such that $L'(s_0)\models \phi$, and
$diff(L'(s_0),L(s_0))$ is minimal;\\
{\tt 05} \hspace*{.2in} return $(M',s_0)$;\\
{\tt 06} \hspace*{.1in} {\bf end}

\vspace*{.1in}

It is easy to observe that this procedure is implemented as the PMA belief update 
\cite{Winslett88}. It is used in the lowest level in our CTL model update prototype.

\vspace*{.1in}
\noindent
{\bf Handling modal formulas $\mbox{AF}\phi$, $\mbox{EX}\phi$ and $\mbox{E}[\phi_1\cup\phi_2]$}

\vspace*{.1in}
From the De Morgan rules and equivalences displayed in Section 2.1, we know that 
all CTL formulas with modal operators can be expressed in terms of these three 
typical CTL modal formulas. Hence it is sufficient to only give the update
functions for these three types of formulas without considering other types of
CTL modal formulas.

\vspace*{.1in}
\noindent
$\ast$ function $\mbox{Update}_{\mbox{AF}}((M,s_0),\mbox{AF}\phi)$ $\ast$\\
input: $(M,s_0)$ and $\mbox{AF}\phi$, where $M=(S,R,L)$, $s_0\in S$, and
$(M,s_0)\not\models\mbox{AF}\phi$;\\
output: $(M',s_0')$, where $M'=(S',R',L')$, $s_0'\in S'$ and 
      $(M',s_0')\models\mbox{AF}\phi$;\\
{\tt 01} \hspace*{.1in} {\bf begin}\\
{\tt 02} \hspace*{.2in}  {\bf if} for all $s\in S$, $(M,s)\not\models \phi$, \\
{\tt 03} \hspace*{.2in} {\bf then} select a state 
$s\in S$ that is reachable from $s_0$, $(M',s^{*})=\mbox{CTLUpdate}((M,s),\phi)$\footnote{Here
$\mbox{CTLUpdate}((M,s),\phi)$ is the main update function that we will describe later.};\\
{\tt 04} \hspace*{.2in} {\bf else} select a path $\pi$ starting from $s_0$ where for all
$s\in\pi$, $(M,s)\not\models \phi$, do (a) or (b):\\
{\tt 05} \hspace*{.4in}  (a) select a state $s\in\pi$, 
$(M',s')=\mbox{CTLUpdate}((M,s),\phi)$;\\
{\tt 06} \hspace*{.4in}  (b)  apply PU2 to disable path $\pi$ and form a new model:\\
{\tt 07} \hspace*{.6in} remove a relation element from $\pi$ that does not affect other paths; \\
{\tt 08} \hspace*{.6in} form a new model $M'=(S',R',L')$:\\  
{\tt 09} \hspace*{.6in} $S'=S$, $R'=R-\{(s_i,s_{i+1})\}$ (note $(s_i,s_{i+1})\subseteq \pi$), and \\
{\tt 10} \hspace*{.6in} $\forall s\in S'$, $L'(s)=L(s)$;\\
{\tt 11} \hspace*{.2in} {\bf if}  $(M',s_0')\models\mbox{AF}\phi$, 
     {\bf then} return $(M',s_0')$\footnote{Here 
$s_0'$ is the corresponding state of $s_0$ in the updated model $M'$, and the same 
for other functions described next.};\\
{\tt 12} \hspace*{.2in} {\bf else} $\mbox{Update}_{\mbox{AF}}((M',s_0'),\mbox{AF}\phi)$;\\
{\tt 13} \hspace*{.1in} {\bf end}

\vspace*{.1in}
Function $\mbox{Update}_{\mbox{AF}}$ handles the update of formula $\mbox{AF}\phi$ as follows:
if no state in the model satisfies formula $\phi$, 
$\mbox{Update}_{\mbox{AF}}$ will first update the model on one state to satisfy $\phi$; 
otherwise, for each path in the model that fails
to satisfy $\mbox{AF}\phi$, $\mbox{Update}_{\mbox{AF}}$ either disables this path in some minimal way, 
or updates this path to make it valid for $\mbox{AF}\phi$. 

\vspace*{.1in}
\noindent
$\ast$ function $\mbox{Update}_{\mbox{EX}}((M,s_0),\mbox{EX}\phi)$ $\ast$\\
input: $(M,s_0)$ and $\mbox{EX}\phi$, where $M=(S,R,L)$, $s_0\in S$, and
$(M,s_0)\not\models\mbox{EX}\phi$;\\
output: $(M',s_0')$, where $M'=(S',R',L')$, $s_0'\in S'$ and
      $(M',s_0')\models\mbox{EX}\phi$;\\
{\tt 01} \hspace*{.1in} {\bf begin}\\
{\tt 02} \hspace*{.2in} do one of (a), (b) and (c):\\
{\tt 03} \hspace*{.4in} (a) apply PU1 to form a new model:\\
{\tt 04} \hspace*{.6in} select a state $s\in S$ such that $(M,s)\models\phi$;\\
{\tt 05} \hspace*{.6in} add a relation element $(s_0,s)$ to form a new model $M'=(S',R',L')$:\\
{\tt 06} \hspace*{.6in} $S'=S$; $R'=R\cup\{(s_0,s)\}$; $\forall s\in S$, $L'(s)=L(s)$;\\
{\tt 07} \hspace*{.4in} (b) select a state $s=succ(s_0)$, 
                                $(M',s^{*})=\mbox{CTLUpdate}((M,s),\phi)$;\\
{\tt 08} \hspace*{.4in} (c) apply PU4 and PU1 to form a new model $M'=(S',R',L')$: \\
{\tt 09} \hspace*{.6in} $S'=S\cup \{s^{*}\}$; $R'=R\cup\{(s_0,s^{*}\}$; $\forall s\in S$,
                 $L'(s)=L(s)$, \\
{\tt 10} \hspace*{.6in}  $L'(s^{*})$ is defined such that $(M',s^{*})\models \phi$;\\
{\tt 11} \hspace*{.2in} {\bf if} $(M',s_0')\models \mbox{EX}\phi$, {\bf then} return
                         $(M',s_0')$;\\
{\tt 12} \hspace*{.2in} {\bf else} $\mbox{Update}_{\mbox{EX}}((M',s_0'),\mbox{EX}\phi)$;\\
{\tt 13} \hspace*{.1in} {\bf end}

\vspace*{.1in}
Function $\mbox{Update}_{\mbox{EX}}$ may be viewed as the implementation algorithm of the
characterization for $\mbox{EX}\phi$ in Theorem 2 in Section 4. However,
it is worth to mentioning that this
algorithm illustrates the difference in $\phi$ in all update
functions from those in the update characterizations and
demonstrates the wider application of the algorithm compared with
their corresponding characterizations. The usage of recursive
calls in the algorithm allows $\phi$ to be an arbitrary CTL formula
rather than a propositional formula as demonstrated in the
characterizations. This is the major difference between the
characterizations and the algorithmic implementation.

\comment{
Function $\mbox{Update}_{\mbox{EX}}$ achieves an update of formula $\mbox{EX}\phi$ 
in such ways similarly to those as described in Theorem 2 in Section 4.2. 
But we should emphasize that Theorem 2 
for characterizing update $\mbox{EX}\phi$ cannot be directly used in function
$\mbox{Update}_{\mbox{EX}}$ because it only considers $\phi$ to be a
propositional formula, while $\phi$ in
function $\mbox{Update}_{\mbox{EX}}$ is an arbitrary
CTL formula.
}

\comment{
in three ways: 
first, if there is a state $s$ in $M$ such that $(M,s)\models\phi$, then it simply adds
a relation element $(s_0,s)$ into $M$ so that the model satisfies $\mbox{EX}\phi$. 
It can also select a successor state $s$ of $s_0$ and update $(M,s)$ to satisfy formula
$\phi$, and consequently the model will then satisfy $\mbox{EX}\phi$. Finally, 
it may add a new successor state $s^{*}$ of $s_0$ such that the new model satisfies
$\phi$ at state $s^{*}$. We should emphasize that Theorem 2 described in Section 4.2
for characterizing update $\mbox{EX}\phi$ cannot be directly used in Function 
$\mbox{Update}_{\mbox{EX}}$ because Theorem 2 only considers $\phi$ to be a 
propositional formula.
}

\vspace*{.1in}
\noindent
$\ast$ function $\mbox{Update}_{\mbox{EU}}((M,s_0),\mbox{E}[\phi_1\mbox{U}\phi_2])$ $\ast$\\
input: $(M,s_0)$ and $\mbox{E}[\phi_1\mbox{U}\phi_2]$, where $M=(S,R,L)$, $s_0\in S$, and
$(M,s_0)\not\models\mbox{E}[\phi_1\mbox{U}\phi_2]$;\\
output: $(M',s_0')$, where $M'=(S',R',L')$, $s_0'\in S'$ and
      $(M',s_0')\models\mbox{E}[\phi_1\mbox{U}\phi_2]$;\\
{\tt 01} \hspace*{.1in} {\bf begin}\\
{\tt 02} \hspace*{.2in}  {\bf if} $(M,s_0)\not\models \phi_1$, {\bf then} 
                          $(M',s_0')=\mbox{CTLUpdate}((M,s_0),\phi_1)$;\\
{\tt 03}  \hspace*{.2in} {\bf else} do (a) or (b):\\
{\tt 04} \hspace*{.4in} (a) {\bf if} $(M,s_0)\models \phi_1$, and 
there is a path $\pi=[s^{*},\cdots]$ ($s_0\neq s^{*}$) \\
{\tt 05} \hspace*{.8in} such that $(M,s^{*})\models \mbox{E}[\phi_1\mbox{U}\phi_2]$,\\
{\tt 06} \hspace*{.6in} {\bf then} apply PU1 to form a new model $M'=(S',R',L')$:\\
{\tt 07} \hspace*{.8in} $S'=S$; $R'=R\cup\{(s_0,s^{*}\}$; $\forall s\in S$ $L'(s)=L(s)$;\\
{\tt 08} \hspace*{.4in} (b) select a path $\pi=[s_0,\cdots,s_i,\cdots,s_j,\cdots]$;\\
{\tt 09} \hspace*{.6in} {\bf if} $\forall s$ $s_0 < s < s_i$, $(M,s)\models\phi_1$, 
$(M,s_j)\models \phi_2$, \\
{\tt 10} \hspace*{.8in} but $\forall s'$ $s_{i+1}<s' < s_{j-1}$, 
$(M,s')\not\models \phi_1\vee\phi_2$ \\
{\tt 11} \hspace*{.6in} {\bf then} apply PU1 to form a new model $M'=(S',R',L')$:\\
{\tt 12} \hspace*{.8in} $S'=S$; $R'=R\cup\{(s_i,s_j)\}$; $\forall s\in S$, $L'(s)=L(s)$;\\
{\tt 13} \hspace*{.6in} {\bf if}  $\forall s$ $s<s_i$, $(M,s)\models\phi_1$,
and $\forall s'$ $s'>s_{i+1}$, $(M,s')\not\models \phi_1\vee\phi_2$, \\
{\tt 14} \hspace*{.6in} {\bf then} apply PU4 to form a new model $M'=(S',R',L')$:\\
{\tt 15} \hspace*{.8in} $S'=S\cup\{s^{*}\}$; $R'=R\cup \{(s_{i-1},s^{*}), (s^{*},s_i)\}$;\\
{\tt 16} \hspace*{.8in} $\forall s\in S$, $L'(s)=L(s)$, $L(s^{*})$ is defined such that
$(M',s^{*})\models\phi_2$;\\
{\tt 17} \hspace*{.2in} {\bf if} $(M',s_0')\models \mbox{E}[\phi_1\mbox{U}\phi_2]$, 
{\bf then} return $(M',s_0')$;\\
{\tt 18}  \hspace*{.2in} {\bf else} 
$\mbox{Update}_{\mbox{EU}}((M',s_0'),\mbox{E}[\phi_1\mbox{U}\phi_2])$;\\
{\tt 19} \hspace*{.1in} {\bf end}

\vspace*{.1in}

To update $(M,s_0)$ to satisfy formula $\mbox{E}[\phi_1\mbox{U}\phi_2]$, function
$\mbox{Update}_{\mbox{EU}}$ first checks whether $M$ satisfies
$\phi_1$ at the initial state $s_0$. If it does not, 
then $\mbox{Update}_{\mbox{EU}}$  will update this initial state so that
the model satisfies $\phi_1$ at its initial state. This will make 
the later update possible. Then under the condition that
$(M,s_0)$ satisfies $\phi_1$, $\mbox{Update}_{\mbox{EU}}$  
considers two cases: if there is a valid
path in $M$ for formula $\mbox{E}[\phi_1\mbox{U}\phi_2]$, then it simply
links the initial state $s_0$ to this path and forms a new path that 
satisfies $\mbox{E}[\phi_1\mbox{U}\phi_2]$ (i.e. case (a));
or $\mbox{Update}_{\mbox{EU}}$  directly selects a path
to make it satisfy formula $\mbox{E}[\phi_1\mbox{U}\phi_2]$ (i.e. case (b)).

\vspace*{.1in}
\noindent
{\bf Handling logical connectives $\neg$, $\vee$ and $\wedge$}

\vspace*{.1in}
Having the De Morgan rules and equivalences on CTL modal formulas, 
an update for formula $\neg \phi$ can be handled quite easily. In fact
we only need to consider a few
primary forms of negative formulas in our algorithm implementation.
Update on a disjunctive formula
$\phi_1\vee\phi_2$, on the other hand,
is simply implemented by calling $\mbox{CTLUpdate}((M,s_0),\phi_1)$ or
$\mbox{CTLUpdate}((M,s_0),\phi_2)$ in a nondeterministic way. Hence  here we
only describe the function of updating for conjunctive formula $\phi_1\wedge\phi_2$.

\vspace*{.1in}
\noindent
$\ast$ function $\mbox{Update}_{\wedge}((M,s_0),\phi_1\wedge\phi_2)$ $\ast$\\
input: $(M,s_0)$ and $\phi_1\wedge \phi_2$, where $M=(S,R,L)$, $s_0\in S$, and
$(M,s_0)\not\models \phi_1\wedge\phi_2$;\\
output: $(M',s_0')$, where $M'=(S',R',L')$, $s_0'\in S'$ and
      $(M',s_0')\models \phi_1\wedge\phi_2$;\\
{\tt 01} \hspace*{.1in} {\bf begin}\\
{\tt 02} \hspace*{.2in} {\bf if} $\phi_1\wedge\phi_2$ is a propositional formula, 
{\bf then} $(M',s_0')=\mbox{Update}_{prop}((M,s_0),\phi_1\wedge\phi_2)$;\\
{\tt 03} \hspace*{.2in} {\bf else} $(M^{*}, s_0^{*})=\mbox{CTLUpdate}((M,s_0),\phi_1)$;\\
{\tt 04} \hspace*{.5in} $(M',s_0')=\mbox{CTLUpdate}((M^{*},s_0^{*}),\phi_2)$ 
                with constraint $\phi_1$;\\
{\tt 05} \hspace*{.2in} return $(M',s_{0}')$;\\
{\tt 06} \hspace*{.1in} {\bf end}

\vspace*{.1in}
Function $\mbox{Update}_{\wedge}$ handles update for a conjunctive formula in 
an obvious way. Line {\tt 04} indicates that when we conduct the update with $\phi_2$, 
we should view $\phi_1$ as a constraint that the update has to obey. Without this 
condition, the result of updating $\phi_2$ may violate the satisfaction of
$\phi_1$ that is achieved in the previous update. We will address this point in more details
in next subsection.


Finally, we describe the CTL model update algorithm as follows.

\vspace*{.1in}
\noindent
$\ast$ algorithm $\mbox{CTLUpdate}((M,s_0),\phi)$ $\ast$\\
input: $(M,s_0)$ and $\phi$, where $M=(S,R,L)$, $s_0\in S$, and
$(M,s_0)\not\models \phi$;\\
output: $(M',s_0')$, where $M'=(S',R',L')$, $s_0'\in S'$ and
      $(M',s_0')\models \phi$;\\
{\tt 01} \hspace*{.1in} {\bf begin}\\
{\tt 02} \hspace*{.2in} {\bf case}\\
{\tt 03} \hspace*{.4in} $\phi$ is a propositional formula: 
       return $\mbox{Update}_{prop}((M,s_0),\phi)$;\\
{\tt 04} \hspace*{.4in} $\phi$ is $\phi_1\wedge\phi_2$: 
                 return $\mbox{Update}_{\wedge}((M,s_0),\phi_1\wedge\phi_2)$;\\
{\tt 05} \hspace*{.4in} $\phi$ is $\phi_1\vee\phi_2$:
return $\mbox{Update}_{\vee}((M,s_0),\phi_1\vee\phi_2)$;\\
{\tt 06} \hspace*{.4in} $\phi$ is $\neg\phi_1$: 
return $\mbox{Update}_{\neg}((M,s_0),\neg \phi_1)$;\\
{\tt 07} \hspace*{.4in} $\phi$ is $\mbox{AX}\phi_1$: return 
$\mbox{CTLUpdate}((M,s_0),\neg\mbox{EX}\neg\phi_1)$;\\
{\tt 08} \hspace*{.4in} $\phi$ is $\mbox{EX}\phi_1$: return
$\mbox{Update}_{\mbox{EX}}((M,s_0),\mbox{EX}\phi_1)$;\\
{\tt 09} \hspace*{.4in} $\phi$ is $\mbox{A}[\phi_1\mbox{U}\phi_2]$: return
$\mbox{CTLUpdate}((M,s_0), \neg (\mbox{E}[\neg\phi_2\mbox{U} (\neg \phi_1\wedge\phi_2)]\vee
      \mbox{EG}\neg\phi_2))$;\\
{\tt 10} \hspace*{.4in} $\phi$ is $\mbox{E}[\phi_1\mbox{U}\phi_2]$: return
$\mbox{Update}_{\mbox{EU}}((M,s_0),\mbox{E}[\phi_1\mbox{U}\phi_2])$;\\
{\tt 11} \hspace*{.4in} $\phi$ is $\mbox{EF}\phi_1$; return
$\mbox{CTLUpdate}((M,s_0),\mbox{E}[\top\mbox{U}\phi_1])$;\\
{\tt 12} \hspace*{.4in} $\phi$ is $\mbox{EG}\phi_1$: return
$\mbox{CTLUpdate}((M,s_0),\neg\mbox{AF}\neg\phi_1)$;\\
{\tt 13} \hspace*{.4in} $\phi$ is $\mbox{AF}\phi_1$: return
$\mbox{Update}_{\mbox{AF}}((M,s_0),\mbox{AF}\phi_1)$;\\
{\tt 14} \hspace*{.4in} $\phi$ is $\mbox{AG}\phi_1$: return
$\mbox{CTLUpdate}((M,s_0),\neg\mbox{E}[\top\mbox{U} \neg\phi_1])$;\\
{\tt 15} \hspace*{.2in} {\bf end case};\\
{\tt 16} \hspace*{.1in} {\bf end}

\vspace*{.1in}

\begin{theorem}
Given a CTL Kripke model $M=(S,R,L)$ and a satisfiable CTL formula
$\phi$, where $(M,s_0)\not\models\phi$ and $s_0\in S$. 
Algorithm $\mbox{\em CTLUpdate}((M,s_0),\phi)$ terminates and generates
an admissible model to satisfy $\phi$. In the worst case, 
$\mbox{\em CTLUpdate}$ runs in time 
${\mathcal O}(2^{|\phi|}\cdot |\phi|^{2}\cdot (|S|+|R|)^{2})$.
\end{theorem}

\noindent
\begin{proof}
Since we have assumed that $\phi$ is satisfiable, 
from above descriptions, it is not difficult to see that
$\mbox{CTLUpdate}$ will only call these functions finite times, and
each call to these functions will (recursively) generate a result that 
satisfies the underlying updated formula, and then return to the main algorithm 
$\mbox{CTLUpdate}$. So $\mbox{CTLUpdate}((M,s_0),\phi)$ will terminate,
and the output model $(M',s_0')$ satisfies $\phi$. 

We can show that the output model $(M',s_0')$ is admissible by induction on the
structure of $\phi$. The proof is 
quite tedious - it involves detailed examinations on $\phi$ running through
each update function. 
Here it is sufficient to observe that for each update function, each time the input model is 
updated in a minimal way, e.g., it adds one state or relation element, removes a minimal
set of relation elements to disconnect a state, or updates a state
minimally. With iterated updates on sub-formulas of $\phi$, minimal changes on the 
original input model will be retained. 

Now we consider the complexity of $\mbox{CTLUpdate}$.
We first analyze these functions' complexity without considering their embedded
recursions.
Function $\mbox{Update}_{prop}$ is to update a state by a propositional formula, which
has the worst time complexity ${\mathcal O}(2^{|\phi|})$. Function 
$\mbox{Update}_{\mbox{AF}}$ contains the following major computations: (1) finding a
reachable state in $(M,s_0)$; (2) selecting a path in which each state does not satisfy
$\phi$; and (3) checking $(M',s_0')\models \mbox{AF}\phi$. Task
(1) can be achieved by computing a spanning tree of $M$ rooted at $s_0$, which can be
done in time ${\mathcal O}(|R|\cdot log|S|)$ \cite{tree02}. Task (2) can be 
reduced to find a valid path for formula 
$\mbox{AG}\phi$. From Theorem \ref{valid}, this can be done in time
${\mathcal O}(|\phi|\cdot (|S|+|R|)^{2})$. Task (3) has the same complexity of task (2).
So, overall, function $\mbox{Update}_{\mbox{AF}}$ has the complexity
${\mathcal O}(|\phi|\cdot (|S|+|R|)^{2})$. Similarly, we can show that functions
$\mbox{Update}_{\mbox{EX}}$ and $\mbox{Update}_{\mbox{EU}}$ have complexity
${\mathcal O}(|\phi|\cdot (|S|+|R|)^{2})$ and 
${\mathcal O}(|\phi|\cdot (|S|+|R|)^{2}+2^{|\phi|})$ respectively. 
Other functions' complexity are obvious either from their implementations based on
the De Morgan rules and equivalences, or from the calls to other functions 
(i.e. $\mbox{Update}_{\neg}$) or the main algorithm (i.e. $\mbox{Update}_{\wedge}$ and
$\mbox{Update}_{\vee}$).
At most algorithm 
CTLUpdate has $|\phi|$ calls to other functions. Therefore, in the worst
time, CTLUpdate runs in time ${\mathcal O}(2^{|\phi|}\cdot |\phi|^{2}\cdot (|S|+|R|)^{2})$.
\end{proof}

\subsection{Discussions}

It is worth mentioning that
except functions $\mbox{Update}_{prop}$, $\mbox{Update}_{\neg}$ and
$\mbox{Update}_{\wedge}$, all other functions used in 
algorithm CTLUpdate are involved in some nondeterministic
choices. This implies that algorithm CTLUpdate is not syntax independent. In other
words, given a CTL model and two logical equivalent formulas,
updating the model with one formula may generate different admissible models.

In the description of function $\mbox{Update}_{\wedge}$, we have briefly mentioned
the issue of constraints in a CTL model update. 
In general, when we perform a CTL model update,
we usually have to protect some properties that should not be violated by
this update procedure. These properties are usually called {\em domain constraints}.
It is not difficult to modify algorithm CTLUpdate to
cope with this requirement. In particular, suppose ${\mathcal C}$ is the set of 
domain constraints for a system specification $M=(S,R,L)$, and we need to
update $(M,s_0)$ with formula $\phi$, where $s_0\in S$, and ${\mathcal C}\cup\{\phi\}$
is satisfiable.  Then in each function of
CTLUpdate, we simply add a model checking condition on the candidate model
$M'=(S',R',L')$: $(M',s_0')\models {\mathcal C}$ ($s_0'\in S'$). 
The result $(M',s_0')$ is returned from the function 
if it satisfies ${\mathcal C}$. Otherwise, the function will 
look for another candidate model.
Since model checking $(M',s_0')\models {\mathcal C}$ can be done in
time ${\mathcal O}(|{\mathcal C}|\cdot (|S'|+|R'|))$, the modified algorithm does not 
significantly increase the overall complexity. In our implemented system prototype, 
we have integrated a generic constraint checking component as an option to be 
added into our update functions so that domain constraints may be taken into account
when necessary.

In addition to the implementation of the algorithm CTLUpdate, we
have implemented separate update functions for typical CTL formulas
such as $\mbox{EX}\phi$, $\mbox{AG}\phi$, $\mbox{EG}\phi$, $\mbox{AF}\phi$,
$\mbox{EF}\phi$, etc., where $\phi$ is a propositional
formula, based on our characterizations provided in Section 4.2.
These functions  simplify an update procedure when the input formula
does not contain nested CTL temporal operators or can be converted into 
such simplified formula.

\section{Two Case Studies}

In this section, we show two case studies of applications of our CTL model 
update approach for system modifications. 
The two cases have been implemented in the CTL model updater prototype, which is a
simplified compiler. In this prototype, the input is a complete CTL Kripke model and a
CTL formula, and the output is an updated CTL Kripke model which satisfies the input
formula. 

We should indicate that our
prototype contains three major components: parsing, model checking and model update functions.
The prototype first 
parses the input formula and breaks it down into its
atomic subformulas. Then the model checking function checks whether the input formula is 
satisfied in the underlying model. If the formula is not satisfied in the model,
our model checking function will generate all relevant states that violate the input formula.
Consequently, this information will directly be used for the model update function to 
update the model. 

\subsection{The Microwave Oven Example}

\begin{figure}[tbhp]
\begin{center}
\epsfysize = 55 mm
\epsffile{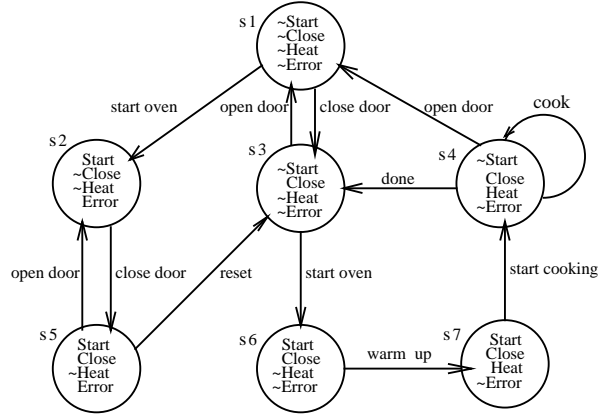}
\caption{CTL Kripke model $M$ of a microwave oven.} 
\label{6.1}
\end{center}
\end{figure}

We consider the well-known microwave
oven scenario presented by Clarke et al. \citeyear{Clarke&etal99}, that
has been used to illustrate the CTL model checking algorithm
on the model describing the behaviour of a microwave oven. 
The Kripke model 
as shown in Figure \ref{6.1} can be viewed as a hardware design of a microwave
oven. In this Kripke model, each state is labeled with both the 
propositional atoms that are true in the state and the negations of propositional
atoms that are false in the state.  The labels on the arcs present
the actions that cause state transitions in the Kripke model. Note
that actions are not part of this Kripke model. The initial state is state 1. 
Then the given Kripke model $M$ describes the behaviour of a microwave oven.

It is observed that this model does
not satisfy a desired property $\phi$ $=$ $\neg \mbox{EF}(Start$ $\wedge$ 
$\mbox{EG}\neg Heat)$: 
``once the microwave oven is started, 
the stuff inside will be eventually heated'' \cite{Clarke&etal99}\footnote{This formula is 
equivalent to $\mbox{AG}(Start\rightarrow \mbox{AF}heat)$.}. That is,
$(M,s_1)\not\models\phi$.
What we would like to do is to apply
our CTL model update prototype to modify this Kripke model
to satisfy property $\phi$. 
As we mentioned earlier, since our prototype combines formula parsing, model checking and
model update together, the update procedure for this case study does not exactly
follow the generic CTL model update algorithm illustrated in Section 6.

\begin{figure}[tbhp]
\begin{center}
\epsfysize = 55 mm
\epsffile{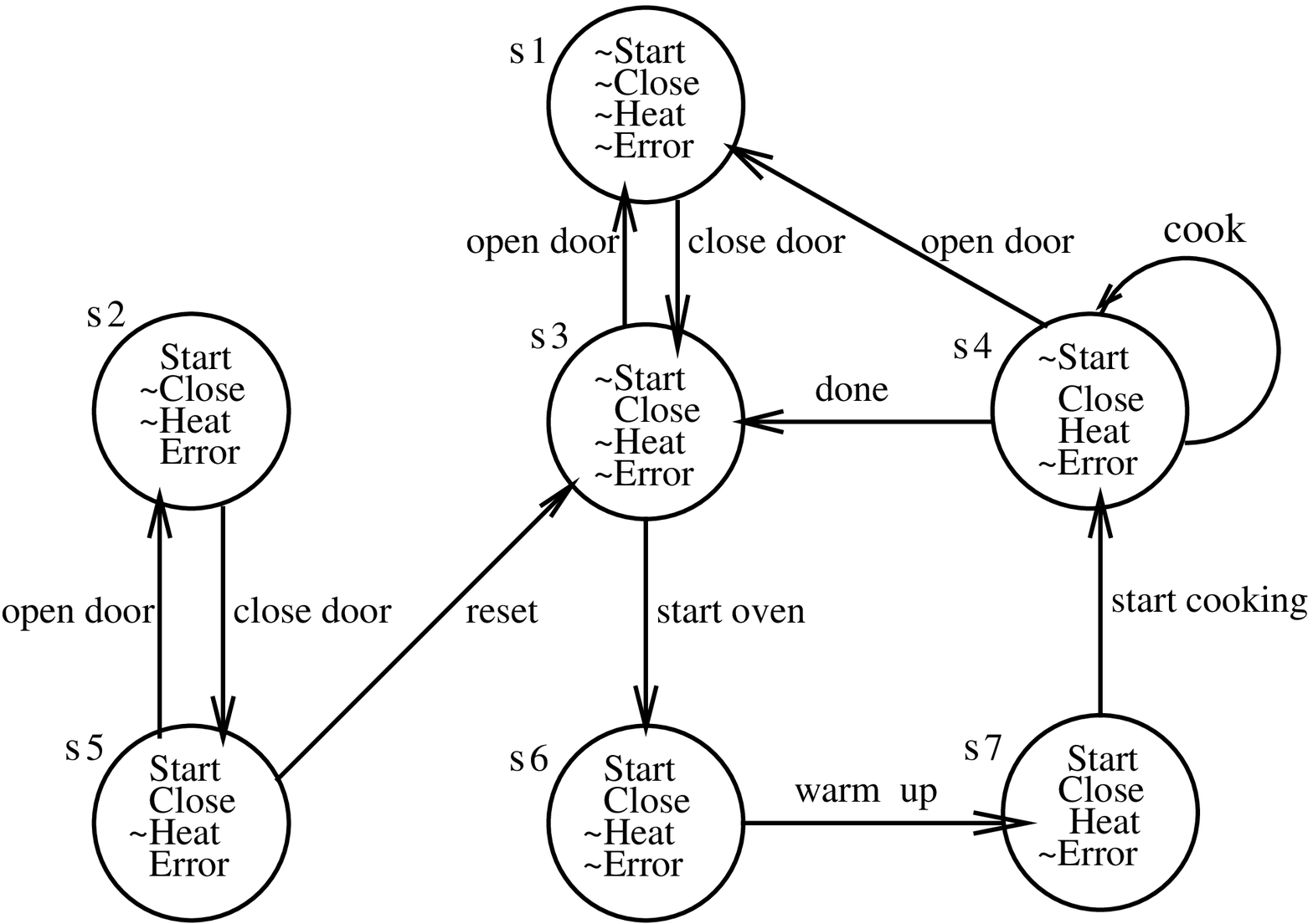}
\caption{Updated microwave oven model using PU2.}
\label{6.2}
\end{center}
\end{figure}

First, we parse $\phi$ into AG$(\neg
(Start\wedge$EG$\neg Heat))$ to remove the front $\neg$. The
translation is performed by function Update$_{\neg}$, which is
called in $\mbox{CTLUpdate}((M,s_1),\phi)$.
Then we check whether each state
satisfies $\neg (Start\wedge$EG$\neg Heat)$.
First, we select EG$\neg Heat$ to be
checked using the model checking function for EG. In this model checking, 
each path that has every state with $\neg Heat$ is identified. Here we find
paths $[s_1, s_2, s_5,s_3,s_1,\cdots]$ and $[s_1,s_3,s_1,\cdots]$
which are strongly connected component loops \cite{Clarke&etal99}
containing all states with $\neg Heat$.
Thus the model satisfies $\mbox{EG}\neg Heat$.
Consequently, we identify all states with $Start$: they are
$\{s_2,s_5,s_6,s_7\}$. Now we select those states with both
$Start$ and $\neg Heat$: they are $\{s_2,s_5\}$. Since the formula
AG$(\neg (Start\wedge$EG$\neg Heat))$ requires that the model
should not have any states with both $Start$ and $\neg Heat$, we
should perform model update related to states $s_2$ and $s_5$.
Now, using Theorem 3 in Section 4.2, the proper update is performed.
Eventually, we obtain two 
possible minimal updates: (1) applying PU2 to remove
relation element $(s_1,s_2)$; or 
(2) applying PU3 to change the truth assignments on $s_2$ and $s_5$. 
After the update, the model satisfies formula $\phi$ and it has a minimal
change from the original model $M$. For instance, by choosing the update
(1) above, we obtain a new Kripke model (as shown
in Figure \ref{6.2}), which simply states that no state transition
from $s_1$ to $s_2$ is allowed,  whereas choosing update (2), we 
obtain a new Kripke model (as shown in Figure \ref{6.4}), which 
says that allowing transition from state $s_1$ to state $s_2$ will cause an error  
that the microwave oven could not start
in $s_2$, and this error message will carry on to its next state $s_5$. 
\comment{
To execute option (3),
our program first converts $\neg (Start\wedge$EG$\neg Heat)$ to
$\neg Start\vee \neg$EG$\neg Heat$, then updates $s_2$ and $s_5$
to satisfy either $\neg Start$ or $\neg$EG$\neg Heat$, using the
main function CTLUpdate($\cal M$,$\phi$) and function
Update$_{\neg}$ in Section~\ref{sec:mainfunction} to deal with
''$\vee$'' and ''$\neg$'' in the formula. In our program, updating
$\neg Start$ is chosen since formula $\neg Start$ is simpler than
$\neg$EG$\neg Heat$. After the update, we will obtain a 
model $(M,s_1)\models\phi$. For instance, by choosing the update
operation (1) above, we will obtain a new Kripke model (as shown
in Figure \ref{6.2}), which simply states that no state transition
from $s_1$ to $s_2$ is allowed, whereas applying operations (2)
and (3) will produce two different models (see Figures
\ref{6.3} and \ref{6.4}).
}

\comment{

\begin{figure}[tbhp]
\begin{center}
\epsfysize = 55 mm
\epsffile{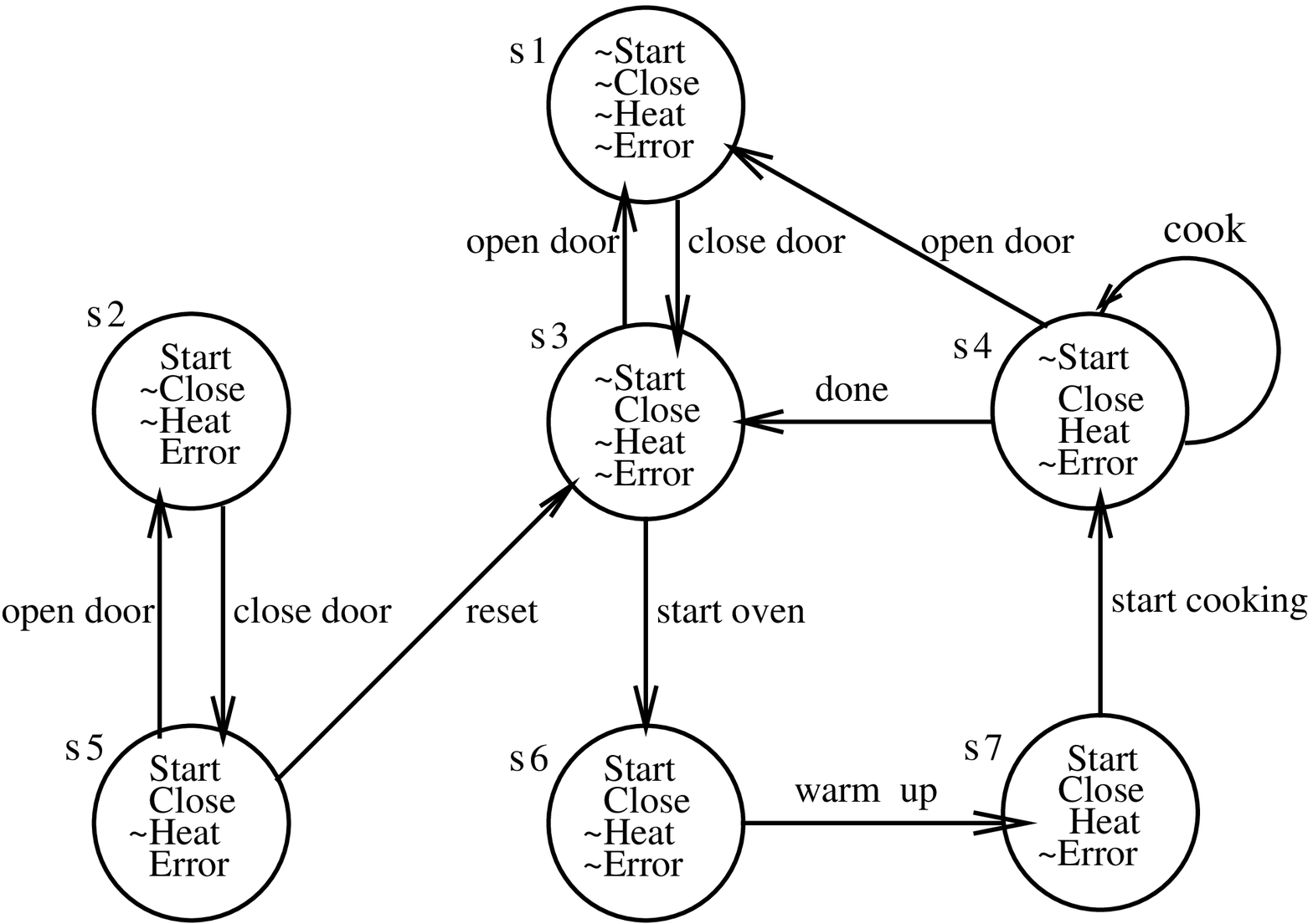}
\caption{Updated microwave oven model using PU2.} 
\label{6.2}
\end{center}
\end{figure}

}


\begin{figure}[tbhp]
\begin{center}
\epsfysize = 55 mm
\epsffile{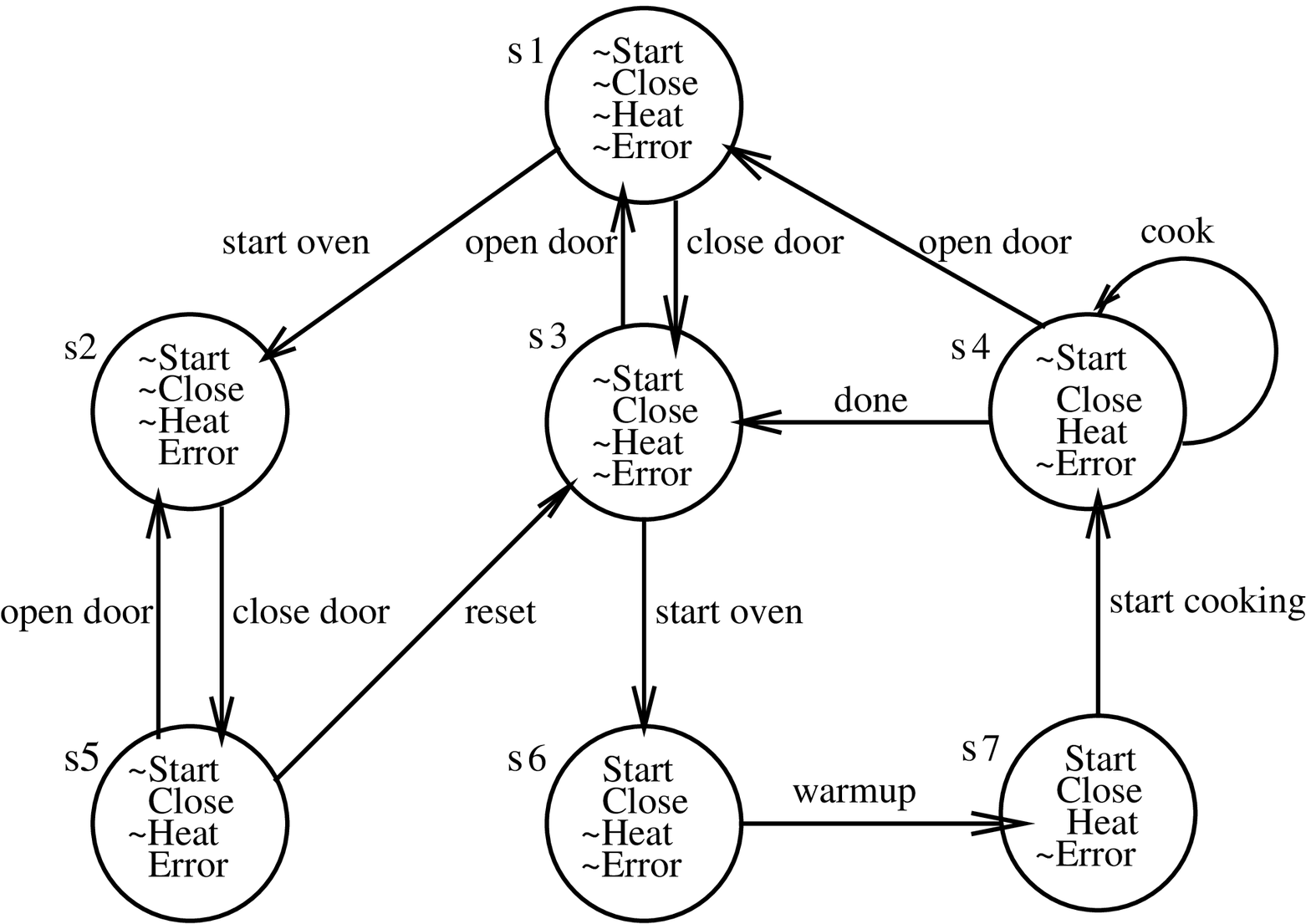}
\caption{Updated microwave oven model using PU3.} 
\label{6.4}
\end{center}
\end{figure}

\comment{
Our algorithm will generate one of the three models
without specific indication because under our criteria they are all
minimally changed from the original model. In our prototype
implementation, however, we have considered the computational cost
of an update upon user request, and its possible impact on 
the system efficiency. For instance, in the above example, with
three possible updates, if the user requires low computational cost,
our system will select the simplest operation (1) to achieve the
goal.
}

\subsection{Updating the Andrew File System 1 Protocol}
\label{sec:AFS1CaseStudy}

\begin{figure}[tbhp]
\begin{center}
\epsfysize = 60 mm
\epsffile{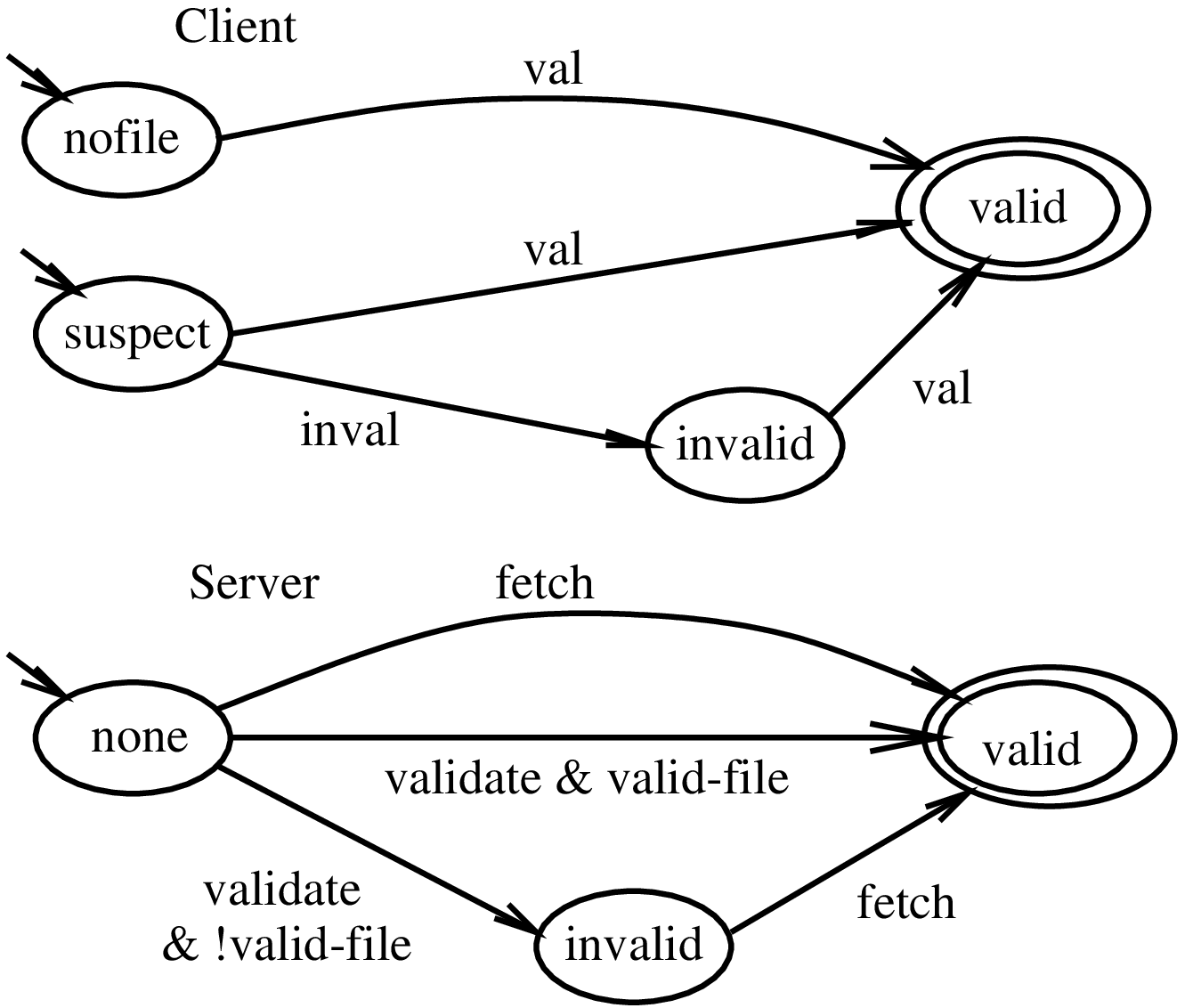}
\caption{State transition diagrams for AFS1.}
\label{fig3}
\end{center}
\end{figure}

The Andrew File System 1 (AFS1)~\cite{wing95} is a cache coherence
protocol for a distributed file system. AFS1 applies a
validation-based technique to the client-server protocol, as
described by Wing and Vaziri-Farahani \citeyear{wing95}. In this protocol, a client has two
initial states: either it has no files or it has one or more files
but no beliefs about their validity. If the protocol starts with the
client having suspect files, then the client may request a file
validation from the server. If the file is invalid, then the client
requests a new copy and the run terminates. If the file is valid,
the protocol simply terminates. AFS1 is abstracted as a model with
one client, one server and one file. The state transition diagrams
with single client and server modules are presented in Figure
\ref{fig3}. The nodes and arcs are labelled with the value for the
state variable, $belief$, and, the name of the received message that
causes the state transition, respectively. A protocol run begins at
an initial state (one of the leftmost nodes) and ends at a final
state (one of the rightmost nodes).

The client's belief about a file has $4$ possible values
$\{nofile, valid, invalid, suspect\}$, where {\em nofile} means that
the client cache is empty; $valid$, if the client believes its
cached file is valid; $invalid$ if it believes its caches file is
not valid; and $suspect$, if it has no belief about the validity of
the file (it could be $valid$ or $invalid$). The server's belief
about the file cached by the client ranges over $\{valid, invalid,
none\}$, where $valid$, if the server believes that the file
cached at the client is valid; $invalid$, if the server believes
it is not valid; {\em none}, if the server has no belief about the
existence of the file in the client's cache or its validity.

The set of messages that the client may send to the server is
$\{fetch, validate\}$. The message $fetch$ stands for a request
for a file, and $validate$ message is used by the client to
determine the validity of the file in its cache. The set of
messages that the server may send to the client is $\{val,
inval\}$. The server sends the $val$ ($inval$) message to indicate
to the client that its cached file is valid (invalid). 
$valid$-$file$ is used when the client has a suspect file
in its cache and requests a validation from the server. If an
update by some other client has occurred then the server reflects
this fact by nondeterministically setting the value of
$valid$-$file$ to $0$; otherwise, $1$ (the file cached at the
client is still valid).
The specification property for AFS1 is:
\begin{equation}
\mbox{AG}((Server.belief=valid)\rightarrow (Client.belief=valid)).
\label{eq:WrongAG}
\end{equation}

In this file system design, the client belief leads the server
belief.
This specification property has been deliberately chosen to fail
with AFS1 \cite{wing95}. Thus, after model updating, we do not need
to pay much attention to the rationality of the updated models.
Our model updater will
update the AFS1 model to derive admissible models which satisfy
the specification property (\ref{eq:WrongAG}).
In this case study, we focus on the update procedure according to 
the functionality of the prototype.

\vspace*{.1in}
\noindent
{\bf Extracting the Kripke model of AFS1 from NuSMV}

\begin{figure}[tbhp]
\begin{center}
\epsfysize = 100 mm
\epsffile{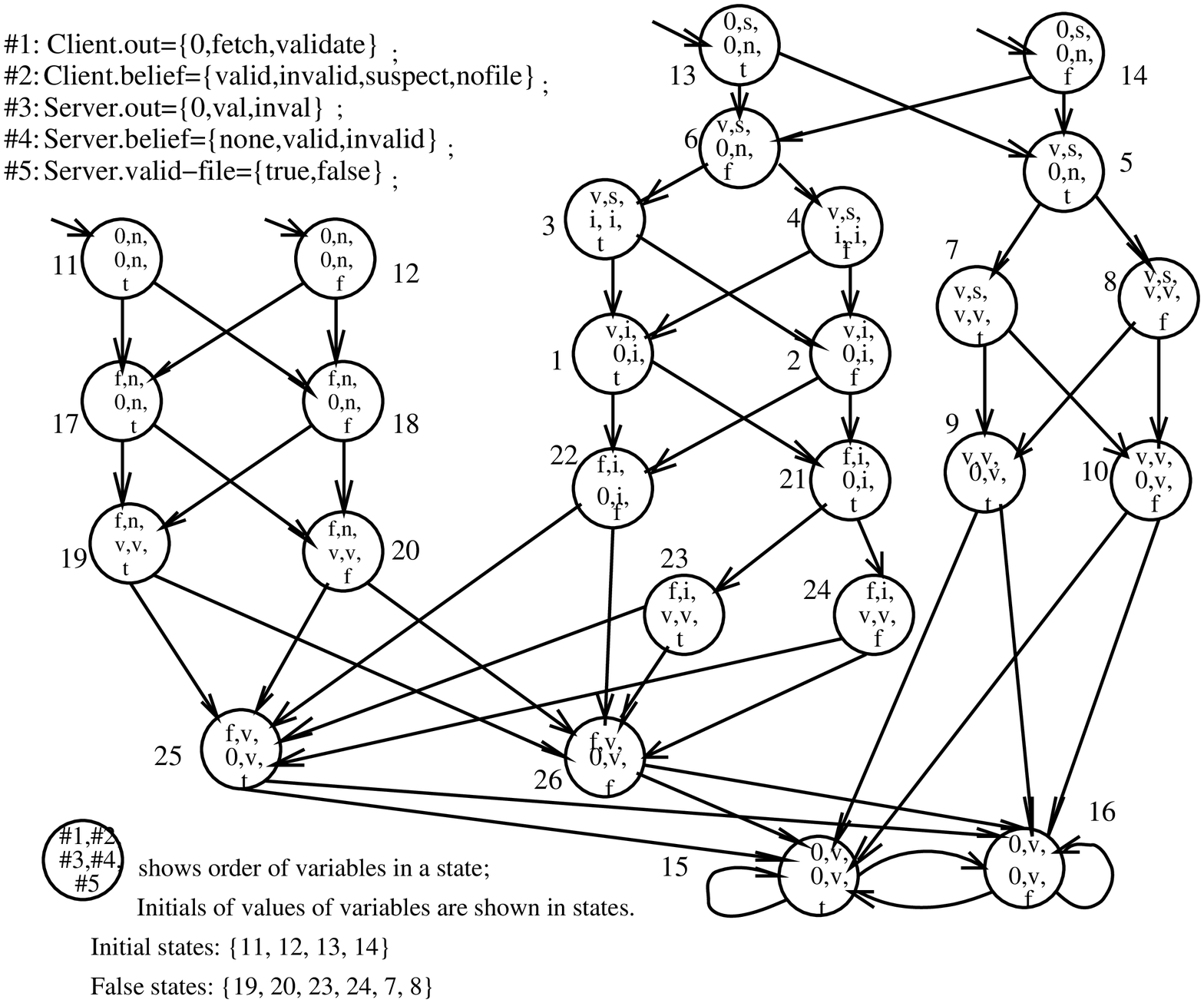}
\caption{CTL Kripke model of AFS1.} 
\label{fig5}
\end{center}
\end{figure}

\vspace*{.1in}
It should be noted that, in our CTL model update algorithm described in
Section~\ref{sec:Algorithms}, the complete Kripke model describing
system behaviours is one of two input parameters (i.e., $(M,s_0)$
and $\phi$), while the original AFS1 model checking process
demonstrated in \cite{wing95} does not contain such a Kripke
model. In fact, it only provides SMV model definitions (e.g.,
AFS1.smv) as input to the SMV model checker.
%
This requires initial extraction of a complete AFS1 Kripke model
before performing any update of it.
For this purpose, 
NuSMV~\cite{Cimatti&etal99} has been used to derive the Kripke
model for the loaded model (AFS1).
The output Kripke model is shown in Figure~\ref{fig5}. This
method can also be used for extracting any other Kripke model.

In the AFS1 Kripke model (see Figure~\ref{fig5}), there are $26$
reachable states (out of total $216$ states) with $52$ transitions
between them. The model contains $4$ initial states
$\{11, 12, 13, 14\}$ and $5$ variables with each individual
variable having $2$, $3$ or $4$ possible values. These variables
are: ``Client.out'', (range $\{0$, $fetch$, $validate\}$);
``Client.belief'' (range $\{valid$, $invalid$, $suspect$,
$nofile\}$); ``Server.out'' (range $\{0$, $val$, $inval\}$);
``Server.belief'' (range $\{none$, $valid$, $invalid\}$); and
``Server.valid-file'' (range $\{true$, $false\}$).

\vspace*{.1in}
\noindent
{\bf Update procedure}

\vspace*{.1in}
\noindent {\em Model checking}: In our CTL model update prototype,
we first check whether formula (\ref{eq:WrongAG}) is
satisfied by the AFS1 model. That is, we need to check whether
each reachable state contains either $Server.belief=\neg valid$ or
$Client.belief=valid$. Our model updater identifies that the set
of reachable states that do not satisfy these conditions is $\{19,
20, 23, 24, 7, 8\}$. We call these states {\em false states}.

\begin{figure}[tbhp]
\begin{center}
\epsfysize = 100 mm
\epsffile{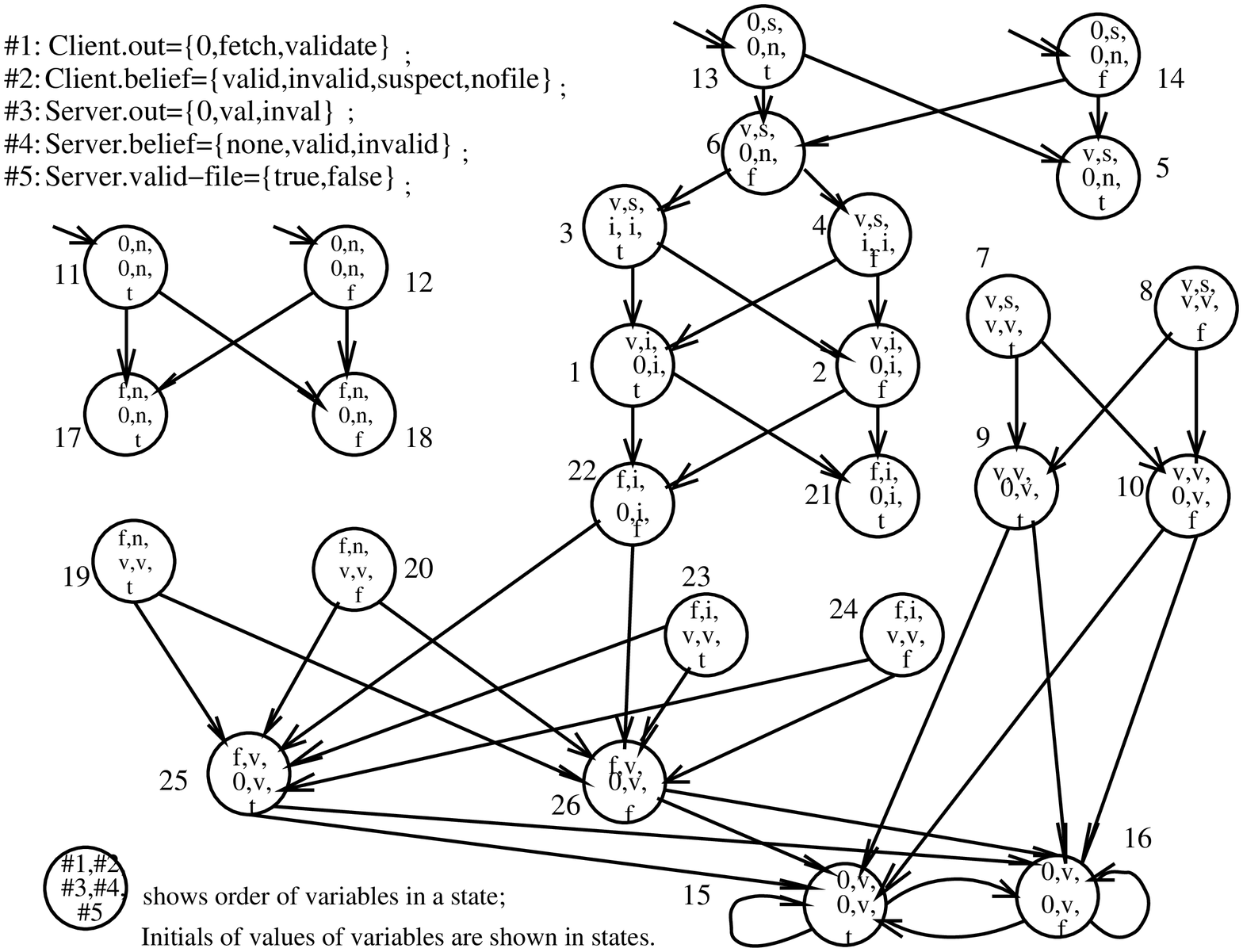}
\caption{One of the admissible models from AFS1 model update.}
\label{figAFS1-new}
\end{center}
\end{figure}

\vspace*{.1in} 
\noindent 
{\em Model update}: Figure \ref{fig5}
reveals that each false state in AFS1 is on a different path.
From Theorem 3 in Section 4 and $\mbox{Update}_{\mbox{AG}}$ in 
Section~\ref{sec:Algorithms}, we know that 
to update the model to satisfy the property,  
operations PU2 and PU3  may be applied to these false states in certain combinations.
\comment{
total number of admissible models from combinations of all
possible updates is $(C^2_1)^6=2^6=64$.
}
As a result, 
one admissible model is depicted in Figure
\ref{figAFS1-new}. 
This model results from the update where each
false state on each false path is updated using PU2.
We observe that after the update, states $25, 26, 15$ and $16$ are
no longer reachable from initial states $11$ and $12$, and states
$9$ and $10$ become unreachable from initial states $13$ and $14$.

We should know that Figure \ref{figAFS1-new} only presents one possible updated
model after the update on AFS1 model. In fact there are too many possible admissible models.
For instance, instead of only using PU2 operation, we could also use both 
PU2 and PU3 in different combinations to produce many other admissible models. 
The total number of such admissible models is 64.

\section{Optimizing Update Results}
\label{ExplosionID}

From Section 7.2, we observe that very often, 
our CTL model update approach may derive many more 
possible admissible models than we really need. In practice, 
we would expect that the solution of a CTL model update provides
more concrete information to correct the underlying system specification.
%
This motivates us to improve our CTL model update
approach so that we can eliminate unnecessary admissible models
and narrow down the update results.

Consider AFS1 update case again. While the model described in Figure 
\ref{figAFS1-new} satisfies the required property and is admissible, it, however, 
does not retain a similar structure to the original 
AFS1 model. This implies that after the update,
there is a significant change to the system behaviour. So this admissible 
model may not represent a desirable correction on the original system. 
One way to reduce this possibility is to impose the notion of {\em maximal reachable states}
into the minimal change principle, so that each possible updated model
will also retain as many reachable states as possible from the original model.

%
Given a Kripke model $M=(S,R,L)$ and $s_0\in S$, and, let ${\mathcal
M}=(M,s_0)$, we say that $s'$ is a {\em reachable state} of
${\mathcal M}$, if there is a path in $M=(S,R,L)$ of the form
$\pi=[s_0,s_1,\cdots]$ where 
$s'\in \pi$. 
$RS({\mathcal M})=RS(M,s_0)$
is used to denote the set of all reachable states of ${\mathcal M}$.
Now, we propose a refined CTL model update principle which can
significantly reduce the number of updated models.
Let $M=(S,R,L)$ be a CTL Kripke model and $s_0\in S$. Suppose
$M'=(S',R',L')$ and $(M',s_0')$ 
is an updated model obtained from the update of $(M,s_0)$ 
to satisfy some CTL formula.  We specify that
\begin{quote}
$RS({\mathcal M})\cap^{\sim} RS({\mathcal M'})=\{s\mid s\in RS({\mathcal M})\cap
RS({\mathcal M'})$ and $L(s)=L'(s)\}$.
\end{quote}
States in $RS({\mathcal M})\cap^{\sim} RS({\mathcal M'})$ are the common reachable states in 
${\mathcal M}$ and ${\mathcal M'}$, called {\em unchanged reachable states}. 
Note that a state having
the same name may be reachable in two different models but with different truth assignments
defined by $L$ and $L'$ respectively. In this case, 
this state is not a common reachable state for ${\mathcal M}$ and ${\mathcal M'}$.

\begin{definition}
\label{def-comm} ({\bf Minimal change with maximal reachable
states}) Given a CTL Kripke model $M=(S,R,L)$, ${\cal
M}=(M,s_{0})$, where $s_{0}\in S$, and a CTL formula $\phi$, a
model $Update({\cal M},\phi)$ is called {\em committed}
with respect to the update of ${\cal M}$ to satisfy $\phi$, if the
following conditions hold: (1) $Update({\cal M},\phi)={\mathcal
M'}=(M',s_0')$ is admissible; and, (2) there is no other 
model ${\mathcal M''}=(M'',s_{0}'')$ such that ${\mathcal
M''}\models \phi$ and $RS({\cal M})\cap^{\sim} RS({\mathcal M'})\subset
RS({\cal M})\cap^{\sim} RS({\mathcal M''})$.
\end{definition}

Condition (2) in Definition \ref{def-comm} ensures that a maximal
set of unchanged reachable states is retained in the updated
model. As we will prove next, the amended CTL model update approach
based on Definition \ref{def-comm} does not significantly increase
the overall computational cost.

\begin{lemma}
Given a CTL Kripke model $M=(S,R,L)$, ${\cal M}=(M,s_{0})$, where
$s_{0}\in S$, a CTL formula $\phi$, and two models
${\cal M'}=(M',s_0')$ and ${\cal M''}=(M'',s_0'')$ from the update of
$(M,s_{0})$ to satisfy $\phi$,
checking whether
$RS({\cal M})\cap^{\sim} RS({\mathcal M'})\subset RS({\cal M})\cap^{\sim}
RS({\mathcal M''})$ can be achieved in polynomial time.
\label{8-1}
\end{lemma}

\noindent
\begin{proof}
For a given $M=(S,R,L)$, we can view $M$ as a directed graph
$G(M)=(S,R)$, where $S$ is the set of vertices and $R$ represents
all edges in the graph. Obviously, the problem of finding all
reachable states from $s_0$ in $M$ is the same as that of finding
all reachable vertices from vertex $s_0$ in graph $G(M)$, which can
be obtained by computing a spanning tree with root $s_0$ in $G(M)$.
It is well known that a spanning tree can be computed in polynomial
time \cite{tree02}. Therefore, all sets $RS({\cal M})$,
$RS({\mathcal M'})$, and $RS({\mathcal M''})$ can be obtained in
polynomial time. Also, $RS({\cal M})\cap^{\sim} RS({\mathcal M'})\subset
RS({\cal M})\cap^{\sim} RS({\mathcal M''})$ can be checked in polynomial
time.
\end{proof}

\begin{theorem}
Given two CTL Kripke models $M=(S,R,L)$ and $M'=(S',R',L')$, where
$s_0\in S$ and $s_0'\in S'$,  and a CTL formula $\phi$, it is
co-NP-complete to decide whether $(M',s_0')$ is a committed result
of the update of $(M,s_{0})$ to satisfy $\phi$.
\label{committed-coNP}
\end{theorem}

\noindent
\begin{proof}
Since every committed result is also an admissible one, from
Theorem \ref{thNP}, the hardness holds. For the membership, we
need to check (1) whether $(M',s_0')$ is admissible; and, (2) an
updated model $M''$ does not exist such that
$(M'',s_0'')\models\phi$ and $RS({\cal M})\cap^{\sim} RS({\mathcal
M'})\subset RS({\cal M})\cap^{\sim} RS({\mathcal M''})$. From Theorem
\ref{thNP}, checking whether $(M',s_0')$ is in co-NP. For (2), we
consider its complement: a updated model $M''$ exits such
that $(M'',s_0'')\models \phi$ and $RS({\cal M})\cap^{\sim} RS({\mathcal
M'})\subset RS({\cal M})\cap^{\sim} RS({\mathcal M''})$. From Lemma
\ref{8-1}, we can conclude that the problem is in NP.
Consequently, the original problem of checking (2) is in co-NP.
\end{proof}

As in Section 4, for many commonly used CTL formulas, we can also
provide useful semantic characterizations to simplify the process of
computing a committed model in an update. Here, we present one such
result for formula AF$\phi$, where $\phi$ is a propositional formula.  
Given a CTL model $M=(S,R,L)$ such that $(M,s_0)\not\models\mbox{AF}\phi$ ($s_0\in S$).
We recall that
$\pi=[s_0,\cdots]$ in $(M,s_0)$ is a {\em valid path} of 
$\mbox{AF}\phi$ if there exists some state $s\in\pi$ and $s>s_0$ such that 
$L(s)\models\phi$; otherwise, $\pi$
is called a {\em false path} of $\mbox{AF}\phi$.

\begin{theorem}
\label{committed-AF} Let $M=(S,R,L)$ be a Kripke model, and
${\cal M}=(M,s_0)\not\models \mbox{\em{AF}}\phi$, where $s_0\in S$ and
$\phi$ is a propositional formula. Let
${\mathcal M'}=Update({\cal M},\mbox{\em{AF}}\phi)$
be a model obtained by the following 1 or 2, then ${\mathcal M'}$ is a committed model.
For each false path $\pi=[s_0,s_1,\cdots]$:
\begin{enumerate}
\item if there is no other false path $\pi'$ sharing any common
state with $\pi$, then PU3 is applied to any state $s\in\pi$
($s>s_0$) to change $s$'s truth assignment such that $L'(s)\models
\phi$ and $Diff(L(s),L'(s))$ is minimal; otherwise, this operation is
only applied to a shared state $s_j$ ($j>0$) in maximum number of
false paths;


\item PU2 is applied to remove relation element $(s_0,s_1)$,
if $s_1$ also occurs in another valid path $\pi'$, where
$\pi'=[s_0,s'_1,\cdots,s'_k,s_1,s'_{k+1},\dots]$ and
there exists some $s_i'$ ($1\leq i\leq k$) such that $L(s_i')\models \phi$.
\end{enumerate}
\end{theorem}

\begin{figure} [tbhp]
\begin{center}
\epsfysize = 60 mm
\epsffile{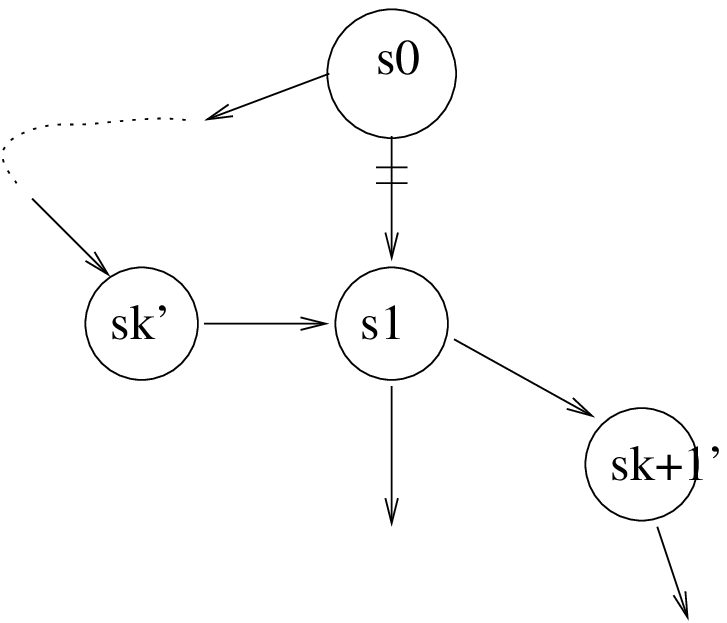}
\caption{$s_1$ occurs in another valid path
$\{s_0,s_1',\cdots,s_k',s_1,s_{k+1}',\cdots]$.}
\label{proof-2}
\end{center}
\end{figure}

\noindent
\begin{proof}
We first prove Result 1. Consider a false path
$\pi=[s_0,\cdots,s_i,s_{i+1},\cdots]$. Since each state in $\pi$
does not satisfy $\phi$, we need to (minimally) change one state $s$'s
truth assignment
along this path so that $L'(s)$ satisfies $\phi$
(i.e., apply PU3 once). If there is no other false path that shares
any states with $\pi$, then we can apply PU3 on any state in path
$\pi$. In this case,
only one reachable state in the original model with respect to this
path is changed to satisfy $\phi$. Thus, the updated model retains
a maximal set of unchanged states.


Suppose that there are other false paths sharing a common state with $\pi$.
Without loss of generality, let 
$\pi'=[s_0,\cdots,s'_{i-1}, s_{i}, s_{i+1}',\cdots]$ be a false path
sharing a common state $s_i$ with $\pi$. Then applying PU3 to any state rather
than $s_i$ in $\pi$ will not necessarily retain a maximal set of
unchanged reachable states, because a further change on any state such as $s_i$
could be made in path $\pi'$ in order to make $\pi'$ valid. Since $s_i$ is a sharing state between
two paths $\pi$ and $\pi'$, it implies that updating two states with PU3 
does not retain a maximal set of
unchanged reachable states comparing to the change only on one state $s_i$ that makes
both $\pi$ and $\pi'$ valid.
\comment{
For example, if a state in $\pi$ which
is not $s_{i}$ has already been changed to make $\pi$
valid, then any change to any state in path $\pi'$ to make $\pi'$
valid will not retain a maximal set of unchanged reachable states
in the original model. However, if we change $s_i$ which is in
both $\pi$ and $\pi'$ to make both paths valid at the same time,
then, the change can retain the maximal set of unchanged reachable
states with respect to path $\pi$ and $\pi'$ after the update.
}

Now we consider the general case.
In order to retain a maximal set of unchanged reachable states in the
original model, we should consider all states in $\pi$ that are also
in other false paths. In this case, we only need to apply PU3 operation
to one state $s_{j}$ in $\pi$ that is shared by a maximal number of
false paths. In this way, changing $s_{j}$ to satisfy $\phi$ will
also 
minimally change other false paths to be valid at the same time.
Consequently, we retain a maximal set of unchanged reachable states
in the original model.

\comment{only need to change one of the common states $\{s_1,\cdots,s_i\}$ in
$\pi$ and $\pi'$ to make both paths valid and, therefore, retain a
maximal set of unchanged reachable states
}

Now we prove Result 2.  Let $\pi=[s_0,s_1,s_2,\cdots]$ be a
false path. According to the condition, there is a valid path
$\pi'$ of the form
$\pi'=[s_0,s'_1,\cdots,s'_k,s_1,s'_{k+1},\cdots]$, where for some
$s'_i\in \pi'$ ($1\leq i\leq k$), $s'_i\models \phi$. Note that
the third path, formed from $\pi$ and $\pi'$,
$\pi''=[s_0,s'_1,\cdots, s'_{k},s_1, s_2,\cdots]$ is also valid.
Applying PU2 on relation element $(s_0,s_1)$
will simply eliminate the false path $\pi$ from the model. Under the
condition, it is easy to see that this operation does not actually
affect the state reachability in the original model because the
valid path $\pi''$ will connect $s_1$ and all states in path $\pi$
are still reachable from $s_0$ but through path $\pi''$.
This is described in Figure \ref{proof-2} as follows.
\end{proof}

\comment{
Different from function $\mbox{Update}_{\mbox{AF}}$ described in Section 6.1, 
Theorem 10 proposes an efficient way to update a CTL model to satisfy 
formula $\mbox{AF}\phi$ where $\phi$ is a propositional formula,
and guarantees that the updated model remains a maximal set of
reachable states from the original model. But this characterization
cannot be directly used for a general update case where we should
consider arbitrary CTL formula that may contain
nested temporal operators.
}

As an optimization of function $\mbox{Update}_{\mbox{AF}}$ described in Section 6.1,
Theorem 10 proposes an efficient way to update a CTL model to
satisfy formula $\mbox{AF}\phi$ to guarantee that the update model retains
a maximal set of reachable states from the original model.
Compared with (a) in function $\mbox{Update}_{\mbox{AF}}$, which updates any state in
a path, case 1 in Theorem 10 only updates a state shared by the
maximum number of false paths to minimize changes in an update to
protect unchanged reachable states. Compared with (b) in function
UpdateAF, which could disconnect a false path to make the
disconnected part unreachable, case 2 in Theorem 10 only
disconnects the false path accompanied by an alternate path to
ensure the disconnected path still reachable via the alternate
path. This theorem illustrates the principle of optimization for
characterizations for other CTL formulas.

\begin{figure} [tbhp]
\begin{center}
\epsfysize = 90 mm
\epsffile{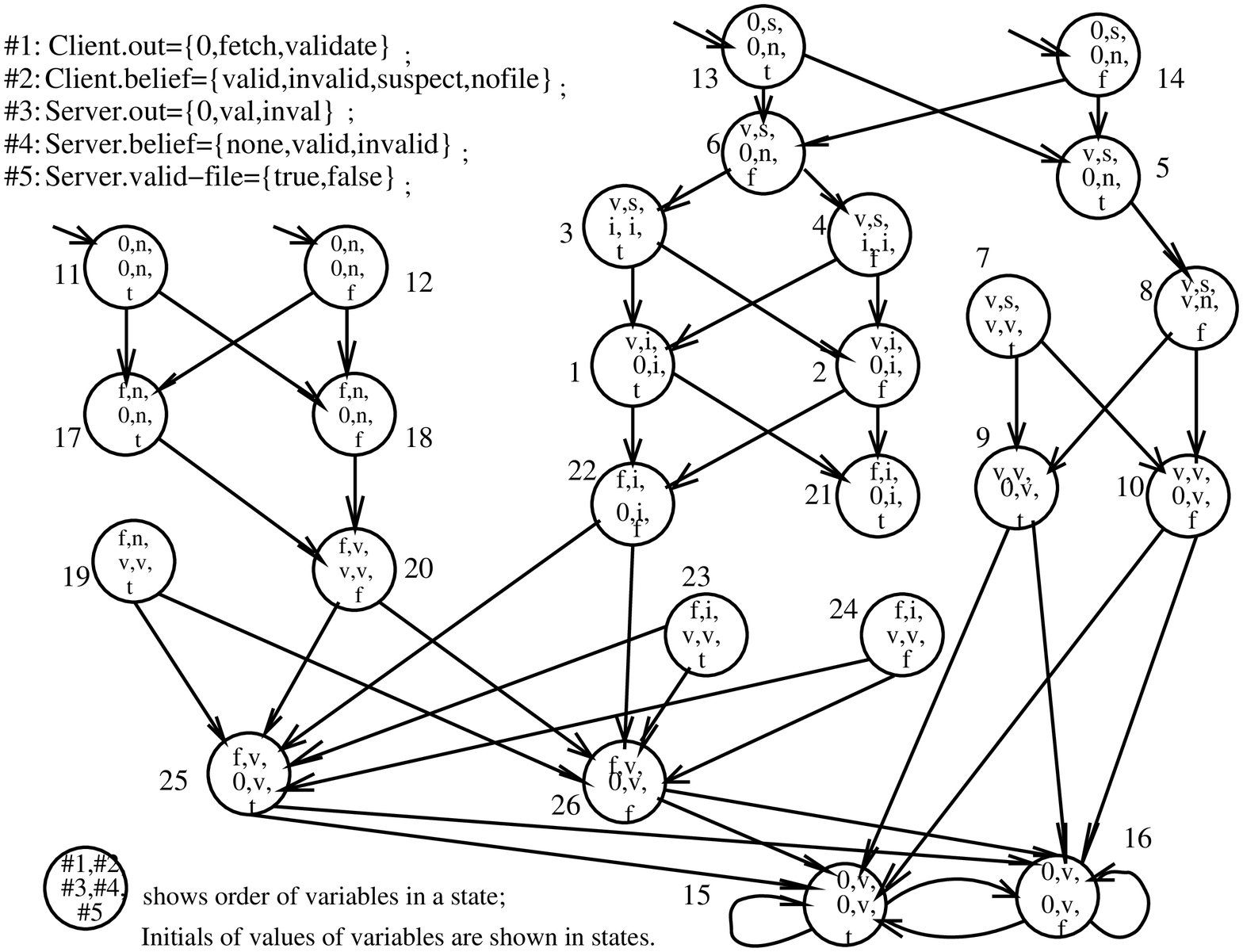}
\caption{One of the committed models of AFS1.}
\label{fig6}
\end{center}
\end{figure}

In general, committed models can be computed by revising our
previous CTL model update algorithms with particular emphasis
to identifying maximal reachable states. 
As an example,
using the improved approach, we can obtain  
a committed model of AFS1 model update (as illustrated in
Figure \ref{fig6}), and rules out the model presented in 
Figure \ref{figAFS1-new}. 
It can be 
shown that using the improved approach to the AFS1 model update, 
the number of total possible updated models is reduced from 64 to 36.

\section{Concluding Remarks}
\label{sec:Conclusions}

In this paper, we present a formal approach to the update of CTL
models. By specifying primitive operations on CTL Kripke
models, we have defined the minimal change criteria for CTL model
update. The semantic and computational properties of our approach
are also investigated in some detail. Based on this formalization,
we have developed a CTL model update algorithm and implemented a
system prototype to perform CTL model update. Two case studies are used
to demonstrate important applications of this work.

There are a number of issues that
merit further investigations. Our current research focuses on the
following two tasks:

\begin{itemize}
\item[-] {\em Partial CTL model update}: In our current approach,
a model update is performed on a complete Kripke model. In
practice, this may not be feasible if the system is complex with a
large number of states and transition relations. One possible
method to handle this problem is to employ the model checker to
extract partial useful information and use it as the model update
input. This could be a counterexample or a partial Kripke model
containing components that should be repaired
\cite{Buccafurri&etal01,Clarke&etal02,Groce&Visser03,Leino&etal03}.
In this way, the update can be directly performed on this
counterexample or partial model to generate possible corrections.
It is possible to develop a unified prototype integrating model
checking (e.g., SMV) and model update.

\item[-] {\em Combining maximal structure similarity with minimal
change}: As demonstrated in Section 8, the principle of minimal
change with maximal reachable states may significantly reduce the
number of updated models. However, it is evident that this
maximal reachable states principle is applied {\em after} the
minimal change (see Definition \ref{def-comm}). We may improve
this principle by defining a unified analogue that integrates both
minimal change and maximal structural similarity at the same
level. This may further restrict the number of final updated
models. This unified principle may be defined based on the notion
of bisimulation of Kripke models \cite{jacm03}. For instance, if
two states are preserved in an update and there is a path between
these two states in the original model, then the new definition
should preserve this path in the updated model as well, so that
the updated model retains maximal structural similarity with
respect to the original. Consider the committed model
described in Figure \ref{fig6}: since there is a path from state
$21$ to state $26$ in the original model (i.e., Figure
\ref{fig5}), we would require retention of the path between $21$
and $26$ in the updated model. Accordingly, the model displayed in
Figure \ref{fig6} should be ruled out as a final updated model.
\end{itemize}

\comment{
Currently, we are in the process
of improving our prototype by integrating it with
the existing model checker.
Note that our current CTL model update system accepts
a complete CTL Kripke model as its input. In practice, this may be inefficient if
we deal with a complex system with a large number of states.  A more
feasible method is to employ a model checker to
extract useful information to use as the model update input. This
could be a counterexample or a partial Kripke model containing
components that should be repaired. In this way, our update can be directly
performed on this counterexample or partial model to generate possible corrections.

}

\section*{Acknowledgments}

This research is supported in part by an Australian Research Council Discovery Grant
(DP0559592). The authors thank three anonymous reviewers for their many valuable 
comments on the earlier version of this paper.

\comment{

}

\bibliographystyle{theapa}
\bibliography{ZhangDing-JAIR2420-Final}

\end{document}